\documentclass[letterpaper, 10 pt, journal]{IEEEtran}

\IEEEoverridecommandlockouts

\usepackage{color}
\usepackage{soul}
\usepackage{CJKutf8}

\title{%
    Memory-Efficient Parallel Policy Learning for Collapse-Reactive Retreat via \\ Hybrid Terrain Modeling
}

\author{Yuki Kadokawa$^{1,2}$, Ritsuki Matsunaga$^{2}$, Yuqi Yang$^{2}$, Edgar Anarossi$^{1}$, Takamitsu Matsubara$^{2}$%
\thanks{
    \textit{(Corresponding author: Yuki Kadokawa.)}
    $^{1}$~Fukushima Institute for Research, Education and Innovation, Namie, Fukushima 979-1521, Japan.
    $^{2}$~Nara Institute of Science and Technology, Ikoma, Nara 630-0192, Japan.
    kadokawa.yuki@naist.ac.jp
}%
}

\usepackage{setspace}
\usepackage{comment}
\usepackage{subfigure}
\usepackage{booktabs}
\ifPDFTeX
  \usepackage{graphicx}
\else
  \usepackage[dvipdfmx]{graphicx} % for no bounding box error
\fi
\usepackage{amsmath}
\usepackage{amsfonts}
\usepackage{algorithmic}
\usepackage{siunitx}
\usepackage{bm}
\usepackage[space, compress]{cite}
\usepackage[ruled,vlined]{algorithm2e}
\usepackage{algorithm2e,setspace}
\usepackage{textcomp}
\usepackage{mathcomp}
\usepackage{tabularx}
\newcolumntype{Y}{>{\raggedright\arraybackslash}X}
\usepackage{here}

\newcommand{\tabref}[1]{Table~\ref{#1}}
\newcommand{\equref}[1]{Eq.~(\ref{#1})}
\newcommand{\figref}[1]{Fig.~\ref{#1}}
\newcommand{\chapref}[1]{Section~\ref{#1}}
\newcommand{\algref}[1]{Algorithm~\ref{#1}}
\newcommand{\apperef}[1]{Appendix~\ref{#1}}

\usepackage{pifont}
\usepackage{placeins}
\providecommand{\cmark}{\ding{51}}
\providecommand{\xmark}{\ding{55}}
\providecommand{\mhead}[2]{\begin{tabular}[b]{@{}c@{}}#1\\\cite{#2}\end{tabular}}

\ifPDFTeX\else\ifXeTeX\else\ifLuaTeX\else\PassOptionsToPackage{dvipdfmx}{hyperref}\fi\fi\fi
\usepackage[colorlinks=true, linkcolor=blue, citecolor=blue, urlcolor=blue]{hyperref}

\renewcommand{\tabref}[1]{\hyperref[#1]{Table~\ref*{#1}}}
\renewcommand{\equref}[1]{\hyperref[#1]{Eq.~(\ref*{#1})}}
\renewcommand{\figref}[1]{\hyperref[#1]{Fig.~\ref*{#1}}}
\renewcommand{\chapref}[1]{\hyperref[#1]{Section~\ref*{#1}}}
\renewcommand{\algref}[1]{\hyperref[#1]{Algorithm~\ref*{#1}}}
\renewcommand{\apperef}[1]{\hyperref[#1]{Appendix~\ref*{#1}}}

\begin{document}

\maketitle

\begin{abstract}
At disaster sites and on unstable slopes, robot contact can collapse uninspected terrain, requiring retreat upon support loss. Massively parallel reinforcement learning of collapse-reactive retreat requires independent state--action experiences. However, on shared rigid-body collapsible terrain, one robot's collapse alters others' contacts without direct interaction. We call this indirect interference. Duplicating terrain objects prevents this but is memory-intensive, limiting parallel sampling. Our key insight is that policy learning does not necessarily require rigid-body collapse modeling, which is feasible but computationally expensive; collapse-induced contact responses can suffice. We therefore propose a memory-efficient parallel policy-learning framework for collapse-reactive retreat via hybrid terrain modeling. Hybrid terrain modeling combines rigid-body simulation of non-collapsible terrain with virtual contact forces for collapsible regions. A State-Multiplexed Contact Model shares terrain geometry and a contact-response law while maintaining environment-specific collapse states, enabling independent experiences without terrain objects for collapsible regions. The framework trains a proprioceptive policy for both normal locomotion and collapse-reactive retreat. It supports over 100 times as many parallel environments as fine-resolution rigid-block terrain within the same GPU memory limit. This scalability and improved sample efficiency enable over 50 times faster policy convergence with comparable retreat performance. Sim-to-sim and real-robot experiments demonstrate retreat under multiple collapse conditions. A supplementary video is available at: \url{https://youtu.be/_RrhwjzEPE0}
\end{abstract}

\begin{figure}[t]
    \centering
    \includegraphics[width=0.99\columnwidth]{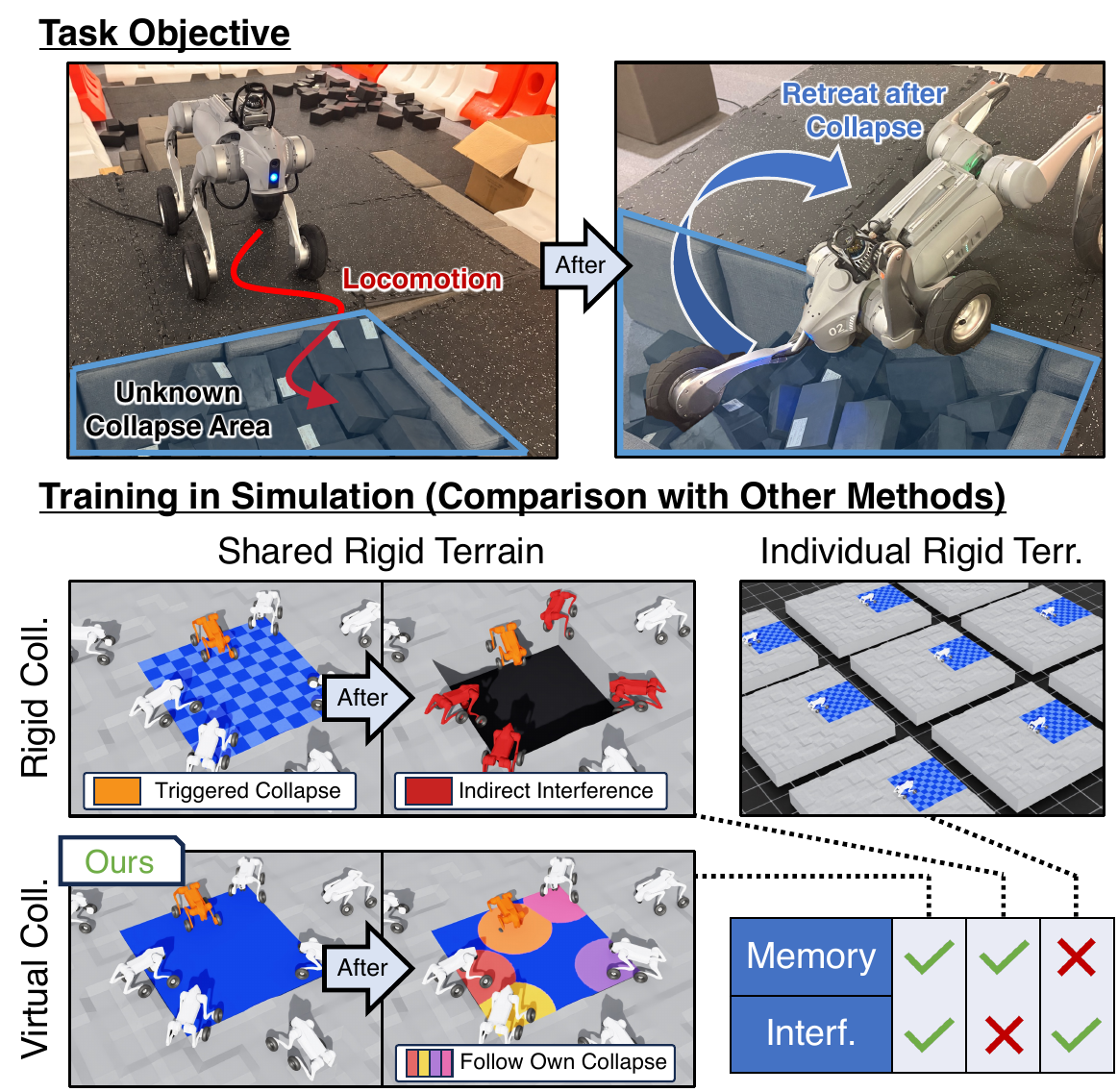}
    \caption{Concept of our policy-learning framework. (Top) A wheeled quadruped encounters collapse-prone terrain that cannot be identified before contact. The objective is to learn a policy that enables the robot to retreat to non-collapsible terrain without falling. (Bottom) Comparison of terrain modeling approaches for parallel policy learning. With rigid collapse, shared rigid terrain enables efficient simulation but causes indirect interference, as collapse induced by one robot affects others. Individual rigid terrain prevents interference but increases simulation cost and training time. With virtual collapse on shared rigid terrain, our hybrid approach combines rigid-body simulation for non-collapsible terrain with virtual contact forces for collapsible regions, enabling memory-efficient parallel learning without indirect interference.}
    \label{fig:promotional_photo}
\end{figure}

\nocite{godon2022mud,godon2025walking,raibert1990trotting,hutter2016anymal,bjelonic2020rolling,he2025residual,haddeler2022traversability,haddeler2022digitaldouble}
\begin{table*}[t]
\centering
\caption{Comparison of Task Strategies for Locomotion on Collapsible Terrain}
\label{tab:rw-strategy}
\footnotesize
\setlength{\tabcolsep}{2pt}
\renewcommand{\arraystretch}{1.15}
\begin{tabular*}{\textwidth}{@{\extracolsep{\fill}}lcccccccccc@{}}
\toprule
Property & \mhead{Perceptive\\RL}{miki2022robust} & \mhead{Vision \&\\Probing}{haddeler2022traversability} & \mhead{SPS}{SPS} & \mhead{Blind\\RL}{lee2020learning} & \mhead{Non-Rigid\\RL}{kim2021nonrigid} & \mhead{Deformable\\RL}{choi2023deformable} & \mhead{RTDD}{haddeler2022digitaldouble} & \mhead{PrePARE}{dey2022prepare} & \mhead{ABS}{he2024agile} & \textbf{Ours} \\
\midrule
(a) Targets ground that collapses under load & \xmark & \cmark & \cmark & \xmark & \cmark & \cmark & \cmark & \xmark & \xmark & \cmark \\
(b) No visual or probing inspection before contact & \xmark & \xmark & \xmark & \cmark & \cmark & \cmark & \cmark & \cmark & \xmark & \cmark \\
(c) Proprioception only, no exteroceptive sensing & \xmark & \xmark & \cmark & \cmark & \xmark & \cmark & \cmark & \cmark & \xmark & \cmark \\
(d) Changes motion after support loss or slip & \xmark & \xmark & \xmark & \xmark & \xmark & \xmark & \xmark & \cmark & \xmark & \cmark \\
(e) Single policy without controller switching & \cmark & \xmark & \xmark & \cmark & \cmark & \cmark & \xmark & \xmark & \xmark & \cmark \\
(f) Validated on a real robot & \cmark & \cmark & \cmark & \cmark & \xmark & \cmark & \cmark & \cmark & \cmark & \cmark \\
\bottomrule
\end{tabular*}
\end{table*}

\def\abstractname{Note to Practitioners}
\begin{abstract}
Robots deployed at disaster sites and on unstable slopes may cross ground that gives way only after it is touched, so that cameras or prior inspection cannot reveal the hazard. This paper presents a way to train, in simulation, a single control policy for a wheeled quadruped that drives normally and, when a wheel loses support, keeps its balance and backs out to firm ground using only its onboard joint and inertial measurements. Training such a policy requires thousands of simulated robots that each experience their own ground collapse. Giving every robot its own copy of a collapsible terrain model quickly exhausts GPU memory, whereas letting the robots share one model allows a collapse caused by one robot to affect the others. Our framework instead represents the collapsible ground by computed contact forces that depend on a small collapse state kept for each robot, while the rest of the terrain is simulated as usual. On a single GPU, this supported over 100 times as many simulated robots as fine rigid-block terrain and reached a converged policy more than 50 times faster, and the trained policy worked without additional training on a real robot in four constructed collapse scenarios. New types of collapse can be added by defining their state and update rules, without building new terrain geometry. Because the policy relies only on proprioception, it may mistake a traversable downward step for a collapse; where such steps are common, combining it with terrain sensing is advisable.
\end{abstract}

\begin{IEEEkeywords}
Terrain modeling, Parallel simulation, Reinforcement learning, Wheeled-legged robots
\end{IEEEkeywords}

\section{Introduction}\label{sec:intro}

    Disaster sites and unstable slopes may contain terrain that subsides or collapses under robot contact, altering support conditions for subsequent contacts~\cite{godon2022mud,godon2025walking}. For legged robots, including wheeled-legged systems, localized loss of support can disturb posture and cause falls~\cite{raibert1990trotting,hutter2016anymal,bjelonic2020rolling,he2025residual}. Collapse risks may not be apparent from visual cues before traversal and may emerge only after contact~\cite{haddeler2022traversability,haddeler2022digitaldouble}. This paper therefore aims to learn a policy for a wheeled quadruped without prior terrain inspection or exteroceptive detection of hazardous regions. Using only proprioceptive observations, the policy performs normal locomotion, responds to collapse-induced posture disturbances and support loss, and retreats to non-collapsible terrain (\figref{fig:promotional_photo}, top).

    Reinforcement learning (RL) is a promising approach to acquiring such a policy by optimizing it with samples from diverse robot--terrain interactions~\cite{hwangbo2019learning,lee2020learning,tan2018sim2real,peng2020learning}. To collect samples efficiently, massively parallel RL is widely used for locomotion-policy learning, simulating many robots simultaneously~\cite{legged_gym,makoviychuk2021isaac,mittal2023orbit}. For fixed terrain, multiple robots can share a single terrain model within the same simulation space while maintaining independent experiences by disabling direct robot--robot interactions~\cite{legged_gym}. Because contact does not alter fixed terrain, one robot cannot change the terrain conditions experienced by others. This sharing avoids per-robot terrain duplication, reducing memory consumption and the computational load on the physics engine.

    However, this terrain-sharing approach cannot be directly applied when robot contact changes the terrain state (i.e., its shape and support conditions). Although rigid bodies can simulate ground collapse~\cite{kim2021nonrigid}, sharing changing terrain causes a region collapsed by one robot to affect the contact conditions and motion of others, even when direct robot--robot interactions are disabled. We refer to this terrain-mediated effect as \emph{indirect interference} (\figref{fig:promotional_photo}, bottom). Assigning independent changing terrain to each robot prevents this interference but requires duplicating terrain objects, i.e., the simulated bodies and colliders that represent the terrain, across parallel environments~\cite{kim2021nonrigid,millard2023granulargym,kadokawa2025prpd}. Finer terrain discretization further increases the number of terrain objects and memory requirements, limiting the number of environments that can run simultaneously~\cite{kadokawa2025prpd}. Thus, the challenge is to achieve computationally efficient parallel policy learning while ensuring that the state--action experiences needed to learn collapse-reactive retreat are generated without indirect interference.

    Our key insight is that, although ground collapse can be simulated using rigid bodies, explicitly modeling the collapsing terrain as rigid bodies is not necessarily required for policy learning. The goal is not to reproduce collapse processes in detail, but to efficiently generate the state--action experiences needed to learn collapse-reactive retreat. We therefore represent collapse through changes in contact responses, particularly support loss and contact forces.

    Based on this insight, we propose a memory-efficient parallel policy-learning framework for collapse-reactive retreat via hybrid terrain modeling. The framework adopts hybrid terrain modeling, which combines rigid-body simulation of the robot and non-collapsible terrain with virtual contact forces for collapsible regions, eliminating the need for terrain objects in these regions. To enable independent robot--terrain interactions within this hybrid approach, we introduce a State-Multiplexed Contact Model (SMCM) that 1) shares reference terrain geometry and a contact-response law across parallel environments, 2) maintains independent collapse states, and 3) computes virtual contact forces according to these states. This allows each robot to experience support loss, subsidence, and lateral slip without indirect interference. Using this environment, we train a proprioceptive policy for both normal locomotion and collapse-reactive retreat through massively parallel RL. Simulation experiments show that the framework supports over 100 times as many parallel environments as fine-resolution rigid-block terrain within the same GPU memory limit. This scalability speeds up sample collection and, together with improved sample efficiency, enables over 50 times faster policy convergence with comparable retreat performance. Sim-to-sim and real-robot experiments further demonstrate successful retreat under multiple tested collapse conditions.

\section{Related Works}

This section reviews two lines of work to which our approach contributes: task strategies for locomotion on collapsible terrain (\chapref{subsec:rw-strategy}, \tabref{tab:rw-strategy}) and simulation methods for state-changing terrain in parallel reinforcement learning (\chapref{subsec:rw-simulation}, \tabref{tab:rw-simulation}).

\begin{table*}[t]
\centering
\caption{Comparison of Simulation Approaches for State-Changing Terrain in Parallel Policy Learning}
\label{tab:rw-simulation}
\footnotesize
\setlength{\tabcolsep}{2pt}
\renewcommand{\arraystretch}{1.15}
\begin{tabular*}{\textwidth}{@{\extracolsep{\fill}}lcccccccc@{}}
\toprule
Property & \mhead{DEM}{cundall1979dem} & \mhead{MPM}{particle_method_sim_meshFree} & \mhead{GranularGym}{millard2023granulargym} & \mhead{PRPD}{kadokawa2025prpd} & \mhead{Movable\\Tiles}{kim2021nonrigid} & \mhead{RFT}{li2013terradynamics} & \mhead{Legged\\Gym}{legged_gym} & \textbf{Ours} \\
\midrule
(a) Terrain state changes with robot contact & \cmark & \cmark & \cmark & \cmark & \cmark & \cmark & \xmark & \cmark \\
(b) Change persists over subsequent contacts & \cmark & \cmark & \cmark & \cmark & \xmark & \xmark & \xmark & \cmark \\
(c) Separate terrain history per environment & \xmark & \xmark & \cmark & \cmark & \xmark & \xmark & \xmark & \cmark \\
(d) No terrain copies per environment & \xmark & \xmark & \xmark & \xmark & \xmark & \cmark & \cmark & \cmark \\
(e) Learning with $\geq 10^{3}$ parallel environments & \xmark & \xmark & \xmark & \xmark & \xmark & \xmark & \cmark & \cmark \\
\bottomrule
\end{tabular*}
\end{table*}

\subsection{Locomotion Strategies for Collapsible and Uncertain Terrain}
\label{subsec:rw-strategy}

Previous works on safe locomotion and navigation over uncertain or hazardous terrain have often formulated the problem as traversability or foothold assessment before contact \cite{kalakrishnan2011learning,fan2021step,yan2026learning}. Vision-based methods predict terrain properties ahead of the robot \cite{wellhausen2019where}, and perceptive locomotion policies combine exteroceptive terrain representations with proprioception \cite{miki2022robust,fang2026motl,luo2026perceptive}. Terrain-probing methods instead assess the ground by physical contact, either with a dedicated probing device at locations selected by vision \cite{haddeler2022traversability} or with the robot's legs \cite{kolvenbach2019haptic,SPS}. In particular, probe-before-step strategies such as the stepping-probing strategy (SPS) test each foothold before transferring body weight onto it and thereby avoid stepping onto ground that may collapse under load \cite{SPS}. These approaches are appropriate when hazardous regions can be identified before traversal, but visual cues may be insufficient for hidden voids or weak support, and repeated probing can substantially limit locomotion speed in time-critical disaster-response tasks \cite{haddeler2022traversability,SPS}.

Other works handle uncertain terrain during traversal rather than before it. Reinforcement learning has produced robust proprioceptive locomotion that transfers from simulation to real robots \cite{hwangbo2019learning,lee2020learning,kumar2021rma}, learned locomotion has been extended to vertically movable tiles \cite{kim2021nonrigid} and granular media \cite{choi2023deformable}, and model-based control has been developed for walking in mud \cite{godon2025walking}. These controllers aim to continue the commanded locomotion on such terrain rather than to stop or retreat after support fails. In other works, hazard handling is added outside the locomotion controller. A real-time digital double (RTDD) detects collapse-related anomalies during walking from discrepancies between the robot and a synchronized physics-based simulation of it \cite{haddeler2022digitaldouble}; predictive proprioception (PrePARE) detects slip events and switches the robot to a more conservative gait \cite{dey2022prepare}; and safety frameworks for legged robots, such as Agile But Safe (ABS), switch from a task policy to a dedicated recovery policy or controller when an unsafe state is predicted or detected \cite{he2024agile,yang2022safe,zhang2025hierarchical}. In these designs, the hazard response relies on a separate detector or switching rule and on whether the activated controller can maintain posture under the hazard.

\tabref{tab:rw-strategy} summarizes these locomotion strategies. Among the closest approaches, Deformable RL~\cite{choi2023deformable} learns a proprioceptive policy for deformable terrain but does not explicitly address retreat after support loss. RTDD~\cite{haddeler2022digitaldouble} detects terrain collapsibility through proprioceptive responses but does not provide a corresponding recovery strategy, while PrePARE~\cite{dey2022prepare} responds to predicted slip events through gait switching rather than learning collapse-reactive retreat. These studies highlight a remaining challenge: learning a unified locomotion policy that not only traverses uncertain terrain but also reacts to unexpected support loss by retreating from the affected region. To address this challenge, we train a single proprioceptive policy for both normal locomotion and collapse-reactive retreat, without prior terrain inspection, explicit hazard detection, or switching to a separate recovery controller.

\subsection{Simulation of State-Changing Terrain for Parallel Reinforcement Learning}
\label{subsec:rw-simulation}

Detailed terrain deformation has been modeled with particle-based and continuum methods, including the material point method (MPM) and the discrete element method (DEM) \cite{particle_method_sim_meshFree,cundall1979dem}. These methods can reproduce local deformation, particle interactions, and history-dependent behavior, and are appropriate for analyzing soil or granular-material mechanics in detail. GPU-based granular simulation (GranularGym) \cite{millard2023granulargym} and progressive-resolution policy distillation (PRPD) \cite{kadokawa2025prpd} have brought particle-based simulation of granular materials into policy learning, although the choice of particle resolution still trades physical fidelity against computational efficiency \cite{kadokawa2025prpd}. Terramechanics provides more compact models of vehicle--terrain interaction with deformable ground, including sinkage and shear-induced slip \cite{wong2001theory}; resistive force theory (RFT) predicts the forces on legs moving through granular media \cite{li2013terradynamics}; and a computationally efficient granular-media model has been used to learn quadrupedal locomotion on deformable terrain \cite{choi2023deformable}. Such compact models, however, primarily describe specific robot--terrain interactions and are not necessarily designed to generate diverse, spatially distributed collapse events whose effects persist over subsequent contacts. For mud, whose behavior upon stepping has been characterized experimentally \cite{godon2022mud}, a simulator suitable for training learned controllers has been reported to be unavailable \cite{godon2025walking}.

\begin{figure*}[t]
    \centering
    \includegraphics[width=1.95\columnwidth]{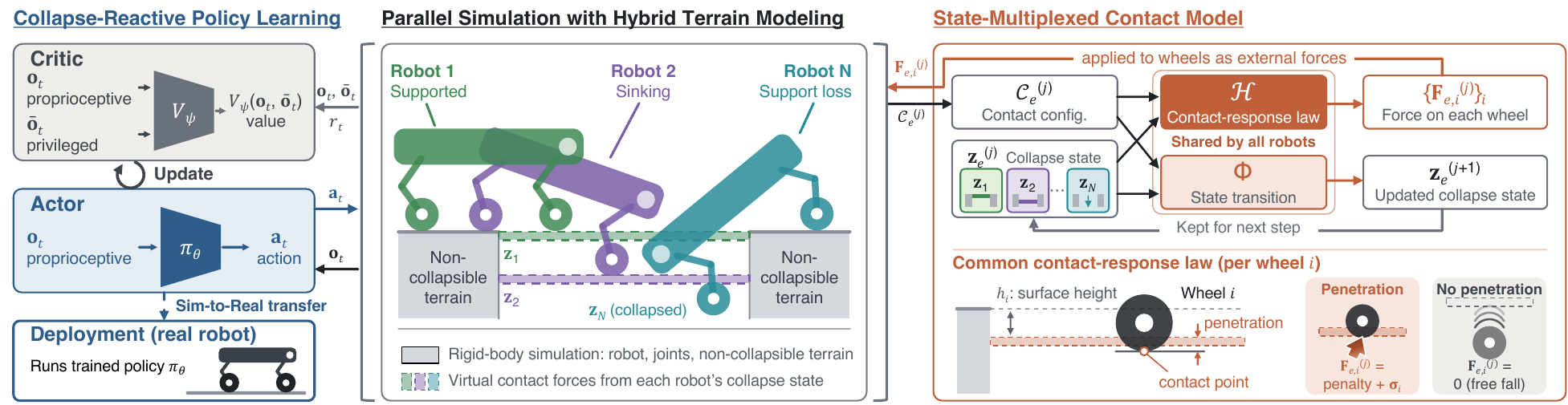}
    \caption{Overview of the proposed policy-learning framework.
\textbf{(Left) Collapse-reactive policy learning:} The actor $\pi_\theta$ receives proprioceptive observations and outputs joint-level references; the critic $V_\psi$ uses both proprioceptive and privileged observations, together with rewards, during training. Only the actor is transferred to the real robot.
\textbf{(Center) Parallel simulation with hybrid terrain modeling:} Samples for policy learning are collected in $N$ parallel environments, in which rigid-body simulation handles the robots and non-collapsible terrain, while the SMCM provides independent virtual contact forces in one shared, collider-free opening. Robots can simultaneously experience supported, sinking, or unsupported terrain without indirect interference.
\textbf{(Right) State-Multiplexed Contact Model:} A shared contact-response law $\mathcal{H}$ maps each wheel contact and its environment-specific collapse state to forces, and the shared transition $\Phi$ updates that state. The reference terrain geometry is also shared. Loss of support allows a wheel to fall through the opening.}
    \label{fig:framework}
\end{figure*}

Simulation-based reinforcement learning is a standard route to locomotion policies \cite{hwangbo2019learning,lee2020learning,tan2018sim2real,peng2020learning}, and GPU-based parallel simulation accelerates it by collecting rollouts from many robots simultaneously \cite{legged_gym,makoviychuk2021isaac,mittal2023orbit}. Locomotion environments commonly build a fixed traversable area from height fields, procedurally generated surfaces, stepping stones, or discrete terrain elements \cite{lee2020learning,legged_gym,RLOC}; because robot contact does not alter this area, thousands of environments can share one terrain mesh without duplicating terrain objects \cite{legged_gym}. For state-changing terrain, however, sharing one changing terrain would let a change caused by one robot alter the contact conditions of others, i.e., cause indirect interference, whereas each rollout must preserve only the history created by its own robot's contacts. When the changing terrain is represented by simulated bodies, such as vertically movable tiles \cite{kim2021nonrigid} or particle beds \cite{millard2023granulargym,kadokawa2025prpd}, keeping the histories independent requires a separate copy in every environment, so terrain memory, contact detection, and state-update costs scale with the number of environments, and GPU memory has been reported to limit the number of parallel environments at fine particle resolutions \cite{kadokawa2025prpd}. The Grid and Particle representations in our experiments follow this per-environment design with rigid blocks and with particles, respectively (\chapref{subsec:exp-baselines}).

\tabref{tab:rw-simulation} summarizes these simulation approaches. Conventional rigid-body and particle-based methods can represent contact-induced terrain changes, but typically require separate terrain objects for each parallel environment to maintain independent interaction histories. In contrast, shared fixed terrain and compact contact models reduce simulation costs but keep no persistent, contact-induced terrain history. This highlights a key challenge for massively parallel RL: representing persistent, contact-induced terrain changes independently across environments without replicating the underlying terrain objects. To address this challenge, we introduce the SMCM, which shares reference terrain geometry and a contact-response law while maintaining independent collapse states for each environment. This design replaces per-environment terrain bodies with compact state representations and virtual contact forces, enabling efficient parallel policy learning (\chapref{subsec:contact-primitive}).

\section{Proposed Framework}
We propose a memory-efficient parallel policy-learning framework (\figref{fig:framework}) that trains a single proprioceptive policy for normal locomotion and collapse-reactive retreat by massively parallel RL on terrain whose state changes after robot contact. The policy acquires collapse-reactive retreat by repeatedly experiencing support loss, subsidence, and lateral slip in thousands of parallel environments, each of which must maintain an independent collapse state determined only by its own robot's contacts. To provide these experiences without creating terrain objects for the collapsible region in every environment, the framework adopts hybrid terrain modeling, which replaces only the collapsible region of the rigid-body simulation with virtual contact forces from a contact model shared by all environments.

The following subsections describe the framework from the overview to the details, in the left-to-right order of \figref{fig:framework}: collapse-reactive policy learning (\chapref{subsec:policy-learning}), the parallel simulation built with hybrid terrain modeling (\chapref{subsec:hybrid-terrain}), and the State-Multiplexed Contact Model (\chapref{subsec:contact-primitive}). Algorithm~\ref{algorithm} gives the pseudocode, and \chapref{sec:experimental_setup} specifies the learning environment used in this work.

\subsection{Collapse-Reactive Policy Learning}\label{subsec:policy-learning}
In each policy update, the current policy collects rollouts in all parallel environments of the simulation (\chapref{subsec:hybrid-terrain}) and is then updated from these rollouts (\figref{fig:framework}, left). The policy is trained with an asymmetric actor--critic architecture~\cite{pinto2018asymmetric}, whose observations are listed in Appendix~\ref{app:observations}. The actor receives only proprioceptive observations available on the real robot, not the collapse-region geometry, collapse state, or distance to the opening, and must infer collapse from contact-induced changes in body motion and joint behavior. The critic additionally receives privileged features of the collapse-capable opening, such as the signed distances of the base and wheels to it, to improve value estimation. During training, a single actor learns both normal locomotion on non-collapsible terrain and collapse-reactive retreat after support loss. At deployment, the critic and all collapse-state variables are discarded; the real robot executes the actor with the same proprioceptive observation and action interfaces used in simulation.

\subsection{Hybrid Terrain Modeling}\label{subsec:hybrid-terrain}
In hybrid terrain modeling (\figref{fig:framework}, center), the rigid-body simulator handles the robot, its actuated joints, and the surrounding non-collapsible terrain of all environments, whereas the contact forces in the collapsible region are computed by the SMCM (\chapref{subsec:contact-primitive}). At each simulation step, the SMCM computes the virtual contact forces of all environments in parallel from the wheel positions, linear and angular velocities, and each environment's collapse state, applies them to the wheels as external forces, and updates the collapse states, after which the rigid-body simulator advances the robot dynamics (Algorithm~\ref{algorithm}). The collapsible region is an actual opening in the surrounding rigid floor, so that a wheel whose virtual support is removed falls through the opening instead of landing on a hidden rigid plane.

This design retains conventional rigid contact on the non-collapsible terrain, where non-penetration is required, whereas no terrain object of the collapsible region is created, deleted, or duplicated for any environment. Because each environment's contact forces are computed from its own collapse state, which only its own robot's contacts update, robots in different environments can simultaneously experience supported, sinking, or unsupported ground at the same location of the opening without indirect interference.

\subsection{State-Multiplexed Contact Model}\label{subsec:contact-primitive}
The SMCM (\figref{fig:framework}, right) separates a collapsible-terrain model into reference terrain geometry, a contact-response law, and a state transition, which are shared by all environments, and a collapse state, which is kept independently for each environment. The contact-response law determines the normal and tangential forces that each robot--terrain contact receives from the virtual ground given the collapse state, and the state transition determines how the contact changes the collapse state. Each concrete collapse phenomenon realized in this way, such as support loss, subsidence, or slip, is called a \emph{collapse primitive}. This separation prevents indirect interference, formally defined in \chapref{subsubsec:shared-contact-law}. Because the form of ground failure is unknown before contact and different forms produce different contact responses, the policy is trained with several collapse primitives so that its retreat does not rely on the contact response of a single form (\chapref{subsec:collapse-primitives}).

\subsubsection{Shared Contact-Response Law and Environment-Specific Collapse State}\label{subsubsec:shared-contact-law}
Let \(e\), \(p\), and \(i\) denote an environment, a collapsible patch, and a contact body, respectively. A patch is a terrain region that the SMCM supports and whose state variables it stores; in this work, each environment has one patch, the collapse-capable opening, which is not subdivided into cells. A contact body is a body of the robot that touches the terrain, i.e., each of the robot's four wheels. The environment-specific collapse state \(\mathbf{z}_e\) collects the state of all patches in environment \(e\): the assigned collapse primitive, its event variables, such as trigger flags, elapsed times, support ratios, subsidence amplitudes, and slip directions, the shape parameters of the collapse region, and the randomized contact and collapse parameters, such as the stiffness of the virtual floor (\chapref{sec:experimental_setup} and \tabref{tab:dr}).

Given the contact-body configuration
\(\mathcal{C}_e=\{\mathbf{x}_{e,i},\mathbf{v}_{e,i},\boldsymbol{\omega}_{e,i}\}_i\), the shared contact-response law \(\mathcal{H}\) and state transition \(\Phi\) are defined as
\begin{equation}
    \{\mathbf{F}_{e,i}^{(j)}\}_i = \mathcal{H}(\mathcal{C}_e^{(j)},\mathbf{z}_e^{(j)}),
    \qquad
    \mathbf{z}_e^{(j+1)} = \Phi(\mathcal{C}_e^{(j)},\mathbf{z}_e^{(j)}),
    \label{eq:state_multiplexing}
\end{equation}
where the superscript \((j)\) denotes the \(j\)-th simulation step of duration \(\Delta t\), \(\mathbf{F}_{e,i}^{(j)}\) is the force with which the virtual ground pushes back on contact body \(i\), and \(\Phi\) updates the event variables in \(\mathbf{z}_e\). Because \(\Phi\) is applied recursively from the initial collapse state \(\mathbf{z}_e^{(0)}\) of an episode, \(\mathbf{z}_e^{(j)}\) summarizes the contact history \(\mathcal{C}_e^{(0:j-1)}=(\mathcal{C}_e^{(0)},\dots,\mathcal{C}_e^{(j-1)})\) of environment \(e\) within the episode. In contrast, if a single changing terrain were shared by all environments, \(\mathbf{z}_e^{(j)}\) would also depend on the contact histories \(\mathcal{C}_{e'}^{(0:j-1)}\) of the other environments \(e'\neq e\); this dependence is the indirect interference that the proposed SMCM eliminates. The superscript \((j)\) is omitted below when it is clear from the context.

For each environment, the SMCM keeps a separate \(\mathbf{z}_e\), which evolves independently and irreversibly within an episode. Adding a parallel environment therefore only adds a few collapse-state variables for the collapsible region, so the number of environments can be increased with little additional memory and computation, which speeds up sample collection.

\subsubsection{Common Contact-Response Law for Collapse Primitives}\label{subsubsec:common-contact-law}
The collapse state \(\mathbf{z}_e\) specifies, for each contact body \(i\), the local surface height \(h_i\), remaining support ratio \(\alpha_i\in[0,1]\), and imposed tangential traction \(\boldsymbol{\sigma}_i\), which are evaluated from the patch states at the position of the contact body. These quantities determine a penalty-based normal response and a tangential response, both gated by the support ratio and by actual penetration, so a wheel that loses support or is above a suddenly lowered surface receives no spurious force and can fall freely. Different collapse primitives are therefore defined by changing only \((h_i,\alpha_i,\boldsymbol{\sigma}_i)\) and the state transition \(\Phi\).

\begin{algorithm}[t]
    \caption{Policy Learning with State-Multiplexed Contact Model}
    \label{algorithm}
    \footnotesize
    Initialize actor parameters \(\theta\), critic parameters \(\psi\), and collapse states \(\{\mathbf{z}_e\}\)\;
    \For{each policy update}{
        Reset completed environments and their collapse states \(\mathbf{z}_e\), and sample their initial states and commands\;
        Initialize rollout buffer \(\mathcal{D}\)\;
        \For{each rollout step \(t\)}{
            Read proprioceptive observation \(\mathbf{o}_t\) and privileged observation \(\bar{\mathbf{o}}_t\)\;
            Sample action \(\mathbf{a}_t\sim\pi_{\theta}(\mathbf{a}_t\mid\mathbf{o}_t)\) and estimate \(V_{\psi}(\mathbf{o}_t,\bar{\mathbf{o}}_t)\)\;
            Apply \(\mathbf{a}_t\) as joint-level references\;
            \For{each simulation step}{
                Read \(\mathcal{C}_e\) for all environments \(e\)\;
                Evaluate \(\{\mathbf{F}_{e,i}\}_i=\mathcal{H}(\mathcal{C}_e,\mathbf{z}_e)\) in parallel\;
                Apply \(\mathbf{F}_{e,i}\) as external forces to the wheels\;
                Update \(\mathbf{z}_e\leftarrow\Phi(\mathcal{C}_e,\mathbf{z}_e)\)\;
                Advance the rigid-body simulator\;
            }
            Compute reward \(r_t\), termination signal \(d_t\), and next observations\;
            Store \((\mathbf{o}_t,\bar{\mathbf{o}}_t,\mathbf{a}_t,r_t,d_t,V_{\psi}(\mathbf{o}_t,\bar{\mathbf{o}}_t))\) in \(\mathcal{D}\)\;
            Update the terrain level for completed environments\;
        }
        Estimate returns and advantages from \(\mathcal{D}\)\;
        Update the asymmetric actor--critic using the collected rollouts\;
    }
\end{algorithm}

\section{Learning Environment Design for Normal Locomotion and Collapse-Reactive Retreat}
\label{sec:experimental_setup}
This section specifies the robot and control interface; three representative collapse primitives, which, together with the common contact-response law, realize \(\mathcal{H}\) and \(\Phi\) in \equref{eq:state_multiplexing}; the surrounding rigid terrain; and the task and learning configuration.

\subsection{Robot and Control Interface}
The target platform is the wheeled quadruped robot Go2W with 12 leg joints and four wheel joints. At every control step \(t\) (50\,Hz), the actor outputs the action \(\mathbf{a}_t\in\mathbb{R}^{16}\) as joint-level references: its scaled entries give position targets for the leg joints, as offsets from the default posture, and velocity targets for the wheel joints. These targets are held over the control step and tracked by joint PD controllers (Appendix~\ref{app:parameters}). This action space lets a single policy combine rolling locomotion with braking, weight redistribution, and backward retreat without switching controllers.

\subsection{Collapse Primitives}\label{subsec:collapse-primitives}
\begin{figure}[t]
    \centering
    \includegraphics[width=\columnwidth]{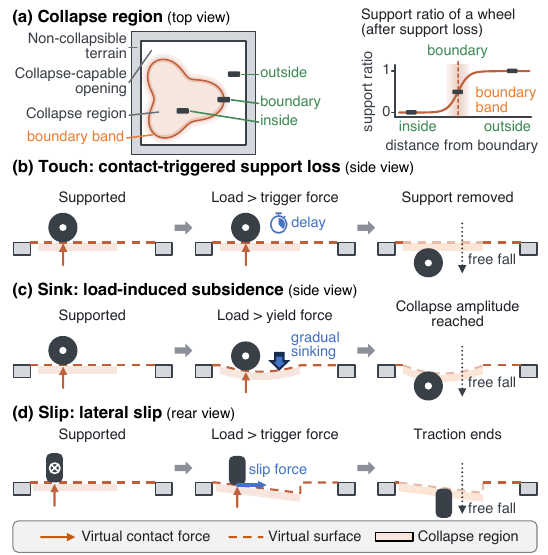}
    \caption{Collapse primitives implemented as representative examples. (a) Collapse-capable opening and a randomly shaped collapse region; the inset shows the support ratio of a wheel across the region boundary after support loss. (b)--(d) Touch, Sink, and Slip, each in three stages from left to right.}
    \label{fig:collapse_primitives}
\end{figure}

The SMCM does not depend on a particular collapse model; a collapse phenomenon enters it only through the collapse state and the state transition (\chapref{subsec:contact-primitive}). This work implements Touch, Sink, and Slip as three representative collapse primitives with qualitatively different contact responses (\figref{fig:collapse_primitives}). Each rule below is one simple realization of its phenomenon, chosen to produce the corresponding contact response; the rules are not physical models of ground failure and are not intended to cover all of its forms. Appendix~\ref{app:collapse} gives their equations, and \tabref{tab:common_params} and \tabref{tab:dr} list their parameters.

All primitives share one contact-response law: a standard penalty-based normal force~\cite{gilardi2002literature,marhefka1999compliant} scaled by the support ratio, regularized Coulomb friction~\cite{castro2020transition}, and an imposed tangential traction. A primitive changes neither this law nor the randomization of its coefficients (\tabref{tab:dr}); through its state transition, it changes only the inputs of the law at each wheel, namely the local surface height, the support ratio, and the imposed tangential traction, which only Slip uses (\chapref{subsubsec:common-contact-law}). Touch and Slip are triggered when a wheel load exceeds a trigger force, whereas Sink progresses while the wheel loads exceed a yield force. All three end with support loss; before it, Touch keeps the surface unchanged for a random delay, Sink lowers the surface, and Slip turns the surface into a depression and imposes a lateral traction. The primitives act on a randomly shaped collapse region inside the collapse-capable opening (\figref{fig:collapse_primitives}(a)).

\subsubsection{Contact-Triggered Support Loss (Touch)}
On irregular terrain such as disaster sites, a foothold can collapse under the load of a robot~\cite{SPS}. Touch represents such a collapse as abrupt support loss: when the normal force on a wheel inside the collapse region first exceeds a trigger force, the region irreversibly loses its support after a random collapse delay (\figref{fig:collapse_primitives}(b)).

\subsubsection{Load-Induced Subsidence (Sink)}
Ground can settle under sustained load and fail when the load exceeds its bearing capacity~\cite{terzaghi1943theoretical}. Sink represents this gradual settlement followed by collapse: the surface of the collapse region sinks while the wheel loads exceed a yield force, and the region loses its support when the subsidence reaches a collapse amplitude (\figref{fig:collapse_primitives}(c)).

\subsubsection{Lateral Slip (Slip)}
Ground can also fail in shear and slide laterally~\cite{terzaghi1943theoretical}. Slip represents a failure of this kind as a lateral push followed by support loss: once triggered as in Touch, the region becomes a depression that deepens along a direction approximately orthogonal to the triggering wheel's velocity; for a short duration, the wheels in contact receive a traction along this direction, and the region then loses its support (\figref{fig:collapse_primitives}(d)).

\subsection{Surrounding Rigid Terrain}
\subsubsection{Terrain Tile and Types}
\begin{figure}[t]
    \centering
    \includegraphics[width=\columnwidth]{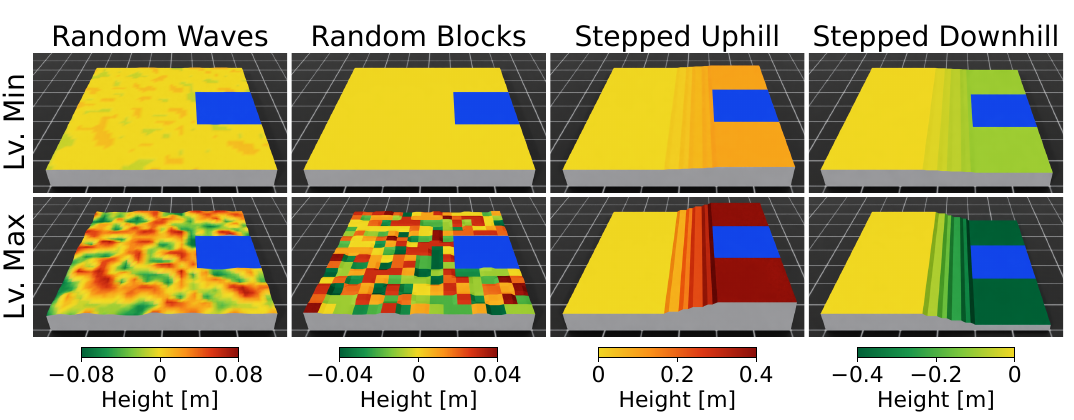}
    \caption{Training terrain types (columns) at the lowest (Lv.~Min, $\ell=0$) and highest (Lv.~Max, $\ell=\ell_{\max}$) terrain levels (rows). Each panel shows a $6\,\mathrm{m}\times6\,\mathrm{m}$ rigid terrain tile; the blue square marks the $2\,\mathrm{m}\times2\,\mathrm{m}$ collapse-capable opening at the middle of the right edge, drawn at its surface height before collapse. Color indicates height; each column's color bar applies to both rows. From Lv.~Min to Lv.~Max, the height amplitude increases from $\pm$0.01 to $\pm$0.08\,m for Random Waves and from 0 to $\pm$0.04\,m for Random Blocks, and the step riser from 0.02 to 0.08\,m for Stepped Uphill and Stepped Downhill.}
    \label{fig:terrain_curriculum}
\end{figure}

Each environment contains a rigid terrain tile with a collapse-capable opening (\figref{fig:terrain_curriculum}). The tile has one of four terrain types: Random Waves (smooth random height undulations), Random Blocks (square blocks with independent random heights), and Stepped Uphill and Stepped Downhill (five steps rising or descending toward the opening, whose nominal surface height \(h_0\) (Appendix~\ref{app:collapse}) is set relative to the top or bottom platform instead of 0\,m). Each type has six terrain levels \(\ell=0,\dots,\ell_{\max}\) (\(\ell_{\max}=5\)), over which the height amplitude or step riser increases linearly.

\subsubsection{Terrain-Mesh Pool}
Each tile uses an exact triangular-mesh collider, because a convex approximation would fill the opening. To avoid costly online collision-mesh changes, each environment holds prebuilt meshes for all type--level combinations; when the terrain level changes (\equref{eq:terrain_curriculum}), the selected mesh is moved into place and the others are parked below the scene.

\subsection{Approach and Retreat Task and Learning Configuration}
\subsubsection{Episode Initialization and Commands}\label{subsubsec:episode-commands}
Each episode starts with the robot on the non-collapsible terrain at a random distance and approach angle from the opening center, facing it. The velocity command begins with a short stop command, after which a forward command drives the robot toward the opening while a yaw-rate command keeps its heading on the opening center. After a wheel enters the opening, a hold phase of random duration keeps this forward command or, in a fraction of episodes, a stop command, during which the robot must avoid falling. A backward command then drives the robot back to its initial position. A stop command of random duration is also inserted once per episode outside the hold phase. \tabref{tab:common_params} lists the corresponding values.

\subsubsection{Reward and Termination Conditions}
Besides command-tracking, stop-command, posture, and regularization terms, the reward includes terms for staying upright after collapse, moving backward and taking the wheels out of the opening under the backward command, completing the retreat, and keeping the center of mass behind the wheels. A fall through the opening or excessive body tilt ends an episode as a failure; a return to the initial position, an exit from the terrain tile, or the time limit truncates it. Appendix~\ref{app:reward} defines all reward terms and termination conditions, and \chapref{subsec:exp-common-settings} defines retreat success.

\subsubsection{Terrain Curriculum}
Following curriculum learning \cite{CRL,zhou2026curriculum} and the terrain curriculum in \cite{legged_gym}, each environment maintains its own terrain level \(\ell\). After each episode, \(\ell\) is promoted on a retreat success (\(\mathrm{succ}\)) and demoted on a failure termination (\(\mathrm{fell}\)) or after repeated episodes in which no wheel enters the opening (\(\mathrm{stuck}\)):
\begin{equation}
    \ell\leftarrow
    \mathrm{clip}
    \left(
        \ell+\mathbf{1}[\mathrm{succ}]
        -\mathbf{1}[\mathrm{fell}\vee\mathrm{stuck}],
        0,\ell_{\max}
    \right).
    \label{eq:terrain_curriculum}
\end{equation}
Each environment keeps its terrain type and collapse primitive across episodes, but one that succeeds at \(\ell_{\max}\) resamples its level and primitive to keep the collapse events diverse. A fixed fraction of the resetting environments also receives a uniformly sampled level so that all levels remain in training (\tabref{tab:common_params}).

\subsubsection{Policy Optimization}
The asymmetric actor--critic is optimized with proximal policy optimization (PPO) \cite{PPO,legged_gym}. For sim-to-real transfer, domain and dynamics randomization~\cite{tobin2017domain,peng2018sim} is applied to the robot's physical properties, the joint PD gains, and the primitives' contact and collapse properties, together with random base pushes and uniform noise on the initial state and actor observations (Appendix~\ref{app:parameters}).
\section{Experiments}
This section first describes the common experimental settings and baselines and then evaluates the proposed framework according to the following research questions.
\begin{list}{}{
  \setlength{\leftmargin}{3.6em}
  \setlength{\labelwidth}{3.1em}
  \setlength{\labelsep}{0.4em}
  \setlength{\itemsep}{0pt}
  \setlength{\parsep}{0pt}
}
  \item[(RQ1)] \textbf{Indirect interference from rigid collapse on shared rigid terrain:} How often does rigid collapse on shared rigid terrain cause indirect interference among parallel environments, and how does this sharing affect the retreat performance of the learned policy compared with individual rigid terrain (\chapref{subsec:exp-interference})?
  \item[(RQ2)] \textbf{Computational efficiency and learning speed:} Compared with individual rigid terrain, does the proposed hybrid terrain modeling improve memory efficiency and simulation throughput (\chapref{subsec:exp-resolution-efficiency}) and thereby reduce the training time required to acquire an effective retreat policy (\chapref{subsec:exp-learning-efficiency})?
  \item[(RQ3)] \textbf{Robustness to collapse primitives and terrain types:} How do the collapse primitives included in training (\chapref{subsec:exp-collapse-ablation}) and the terrain types encountered during evaluation (\chapref{subsec:exp-complex-terrain}) affect the retreat performance of the learned policy?
  \item[(RQ4)] \textbf{Generalization and baseline comparison:} How well does the learned policy generalize to other representations of collapsible terrain (\chapref{subsec:exp-sim2sim}), and how does its retreat performance compare with the baselines, including estimator-triggered retreat (\chapref{subsec:exp-speed-escape-balance})?
  \item[(RQ5)] \textbf{Effect of collapse-exposure design:} How does the direction of motion in which collapse is encountered during training (forward only, as in Ours; backward only; or both) affect retreat performance (\chapref{subsec:exp-policy-limitations})?
  \item[(RQ6)] \textbf{Transfer to physical collapse scenarios:} Can a policy trained in simulation perform collapse-reactive retreat on a real wheeled quadruped in physical collapse scenarios without additional training (\chapref{subsec:exp-real-transfer})?
\end{list}

\subsection{Common Settings}\label{subsec:exp-common-settings}
The simulation environments in the experiments are built on Isaac Lab, and all policies, including those of the baselines, are trained and the computational cost is measured on an NVIDIA GeForce RTX 4090 GPU. All learning-based methods use the same wheeled quadruped robot, Go2W, with the same observation and action spaces and low-level controller, and PPO~\cite{PPO,legged_gym} is used for policy optimization. The training settings of Ours, which are common to all experiments unless otherwise stated, are given in the appendices: the observations in Appendix~\ref{app:observations}, the training parameters and domain randomization in Appendix~\ref{app:parameters}, and the reward function and termination conditions in Appendix~\ref{app:reward}. Policies are trained with a terrain curriculum (\equref{eq:terrain_curriculum}), and \figref{fig:terrain_curriculum} shows the four terrain types used for training. Unless otherwise stated, a retreat is regarded as successful when, after a wheel enters the collapse-capable opening, the base of the robot is at least 0.5\,m from the opening, measured from its near edge, with all four wheels outside the opening and in contact with the ground, without falling.

\subsection{Baselines}\label{subsec:exp-baselines}
We compare Ours with two groups of baselines: terrain-modeling baselines, which simulate the collapsible terrain with rigid bodies instead of hybrid terrain modeling, and method baselines, which replace collapse-reactive retreat with estimator-triggered retreat.

\subsubsection{Terrain-Modeling Baselines}\label{subsubsec:terrain-baselines}
\begin{figure}[t]
    \centering
    \includegraphics[width=0.98\columnwidth]{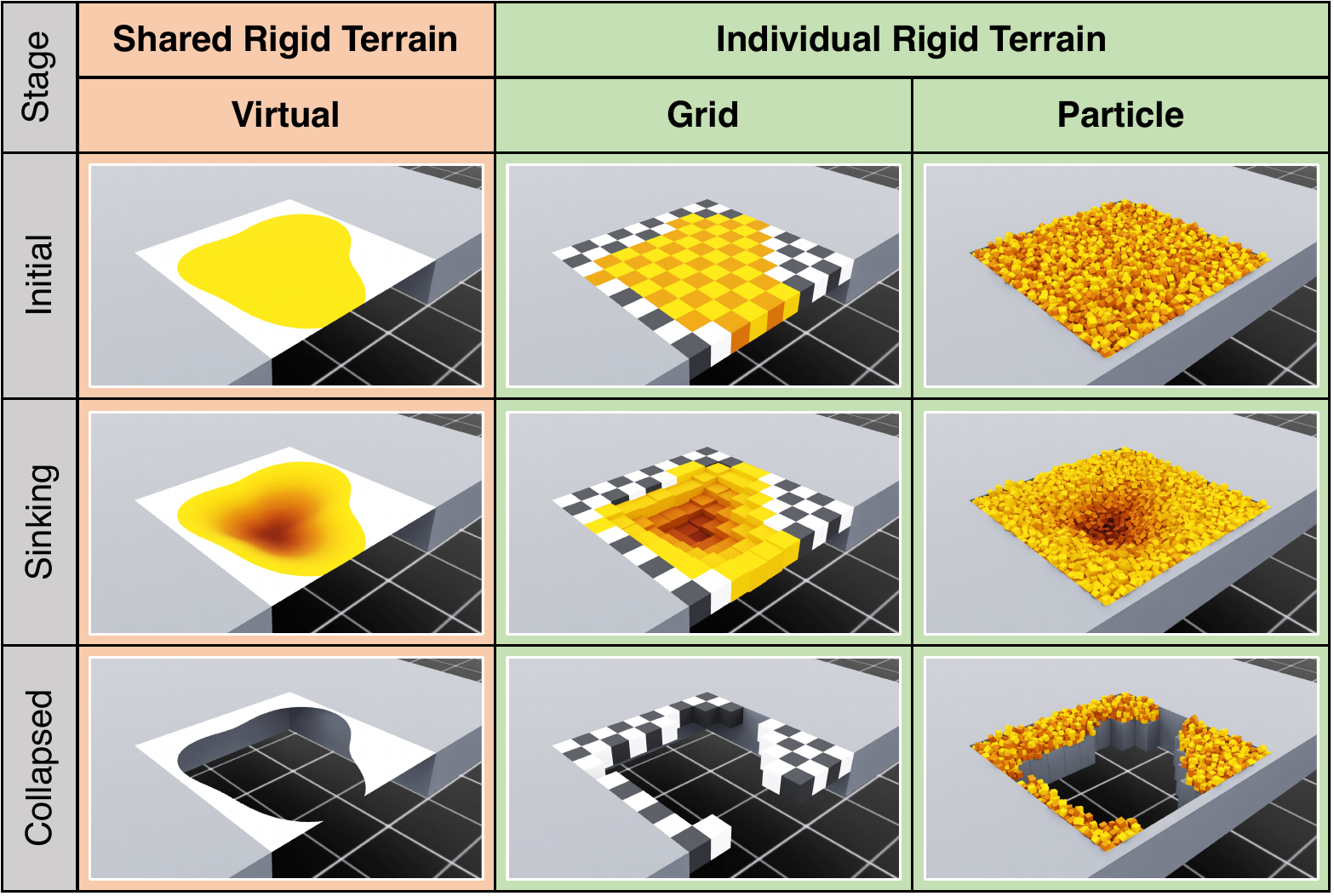}
    \caption{Terrain representations of Ours, Grid, and Particle, illustrated with the Sink collapse primitive. The columns are grouped as in \figref{fig:promotional_photo} (bottom): on shared rigid terrain, Ours (Virtual) models collapse with virtual contact forces (hybrid terrain modeling); on individual rigid terrain, the terrain-modeling baselines Grid (rigid blocks, one per grid cell) and Particle (particles on a floor that is removed cell by cell) simulate collapse with rigid bodies, and each environment has its own copy of these bodies and of their collapse state. From top to bottom, the rows show the Initial, Sinking, and Collapsed stages. Within the $2\,\mathrm{m} \times 2\,\mathrm{m}$ collapse-capable opening in each image, the yellow-to-orange area denotes the collapsing floor. For Ours, the column only illustrates the change in surface shape.}
    \label{fig:sim_view}
\end{figure}

\figref{fig:promotional_photo} (bottom) contrasts three ways of modeling collapsible terrain for parallel policy learning: rigid collapse on shared rigid terrain, rigid collapse on individual rigid terrain, and virtual collapse on shared rigid terrain (the hybrid terrain modeling of Ours). Here, shared and individual indicate whether the parallel environments use one collapsible floor or each have their own copy of its rigid bodies and collapse state; Ours shares the reference terrain geometry of the floor but keeps a virtual collapse state for each environment. For rigid collapse, the collapsible floor can be built from simulated bodies such as vertically movable tiles~\cite{kim2021nonrigid}, rigid blocks, or particles~\cite{millard2023granulargym,kadokawa2025prpd}. We build it from a grid of rigid blocks (Grid), which can represent the collapse regions and collapse primitives used in Ours at a resolution set by the cell size. Grid thus differs from Ours mainly in whether the collapsible floor is simulated by rigid bodies copied for each environment or by virtual contact forces. We compare the following terrain representations (\figref{fig:sim_view}):
\begin{itemize}
    \item \textbf{Grid (individual rigid terrain):} Each environment has its own rigid-block terrain, in which the collapsible floor is divided into rigid blocks, one per grid cell of 50, 33, 20, or 10\,cm (Grid (50)--Grid (10)). The collapse-region shape and collapse primitive are represented at the grid resolution; the blocks are moved by the same collapse-state transitions as Ours, but the wheels are supported by rigid contact with the blocks instead of virtual normal forces. Because the blocks of each environment follow only its own collapse state, this per-environment copy prevents indirect interference at the cost of duplicating terrain objects across environments.
    \item \textbf{Shared Grid (rigid collapse on shared rigid terrain):} The environments in a group share one terrain instance, i.e., the collapse state of one Grid (50) floor, so that a collapse caused by one robot also changes the terrain that the other robots of the group encounter. A changed terrain instance is restored, with a newly drawn collapse primitive and collapse-region shape, when the robot that first changed it is reset; any robot then in the opening is reset with it. It is used only in \chapref{subsec:exp-interference}.
    \item \textbf{Particle (individual rigid terrain):} Each environment has its own particles, distributed across the collapse region and initially supported by a floor, as a more detailed representation than Grid. As in Grid, collapse is triggered on a grid-cell basis: when a cell collapses, its supporting floor is removed, allowing the particles above it to fall and model terrain failure. Because its simulation cost makes policy training impractical, Particle is used only to measure computational cost and as a testing representation in the sim-to-sim experiment (\chapref{subsec:exp-sim2sim}).
\end{itemize}

\subsubsection{Method Baselines}\label{subsubsec:method-baselines}
To compare Ours with other retreat methods, we use two method baselines that pair a locomotion policy trained on non-collapsible terrain in Legged Gym~\cite{lee2020learning,legged_gym} with an estimator of collapse onset; when the estimator detects a collapse, a velocity command in the retreat direction is applied.
\begin{itemize}
    \item \textbf{Locomotion Policy with Naive Fallen Estimator (Naive):} A neural fall estimator receives the same proprioceptive observations as the locomotion policy and is trained by supervised learning on collapsible-terrain rollouts of this policy, with the timing of collapse-induced wheel falls as the label.
    \item \textbf{Real-Time Digital Double (RTDD):} RTDD estimates collapse-related anomalies from discrepancies in body orientation, joint states, and foot forces between the real robot and its synchronized physics-based digital double~\cite{haddeler2022digitaldouble}, both of which are controlled by the locomotion policy.
\end{itemize}

\subsection{RQ1: Indirect Interference from Rigid Collapse on Shared Rigid Terrain}\label{subsec:exp-interference}

\subsubsection{Settings}
To examine how indirect interference caused by rigid collapse on the shared rigid terrain in \figref{fig:promotional_photo} affects policy learning, we train Shared Grid with different numbers of environments per terrain instance and compare its indirect-interference rate during training and sim-to-sim success rate with those of Grid (50), which has one environment per terrain instance.

\subsubsection{Results}
\begin{figure}[t]
    \centering
    \includegraphics[width=0.99\columnwidth]{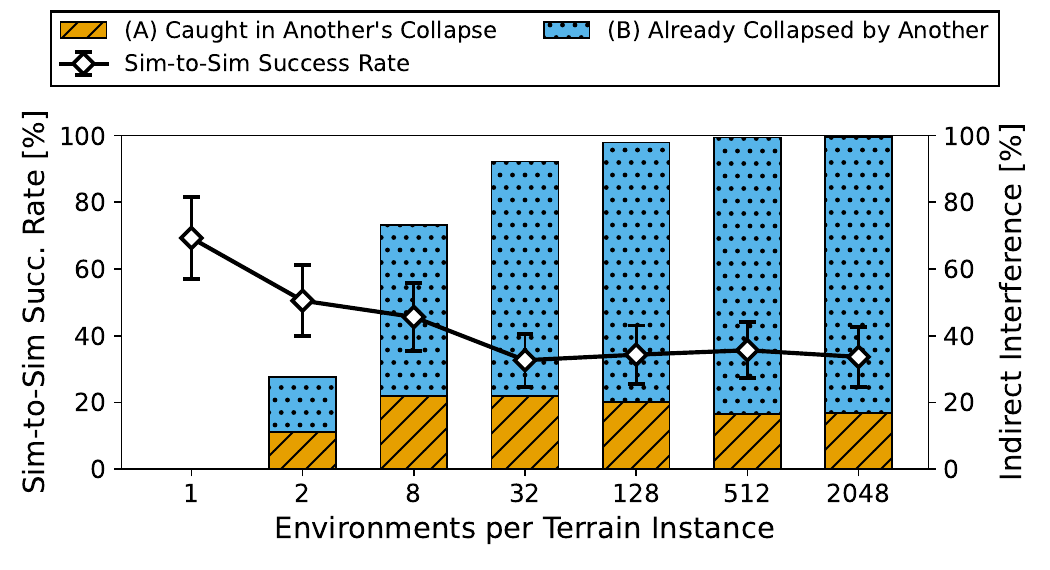}
    \caption{Sim-to-sim success rate (line, left axis) and indirect-interference rate during training (stacked bars, right axis) versus the number of environments per terrain instance among $N=2{,}048$ parallel environments. One environment per terrain instance is Grid~(50), and two or more are Shared Grid. For each setting, five policies are trained independently with the same settings as Grid~(50) in \chapref{subsec:exp-learning-efficiency}. Bars show the mean share of the training episodes reaching the opening in which the robot was first (A) caught in another robot's collapse or (B) on terrain already collapsed by another robot. The success rate is evaluated as in \figref{fig:sim2sim_success_rate}; points and error bars show the mean and standard deviation of its Average column.}
    \label{fig:indirect_interference}
\end{figure}

As shown in \figref{fig:indirect_interference}, indirect interference occurs in about 28\% of the training episodes that reach the opening even with two environments per terrain instance, in about 73\% with eight, and in over 90\% with 32 or more, mostly as type (B). As the interference increases, the sim-to-sim success rate decreases from about 70\% for Grid (50) to about 50\% with two environments per terrain instance, about 45\% with eight, and about 35\% with 32 or more. This suggests that rigid collapse on shared rigid terrain exposes a substantial share of the experiences collected for learning to indirect interference, even when only two environments share a terrain instance, and degrades the learned retreat policy.

\subsection{RQ2: Memory Efficiency and Sample-Collection Speed}\label{subsec:exp-resolution-efficiency}

\subsubsection{Settings}
To evaluate the memory efficiency and sample-collection speed of hybrid terrain modeling, we measure the GPU memory usage and the simulation time per sample of Ours, Grid with cell sizes ranging from coarse to fine, and Particle while increasing the number of parallel environments $N$.

\subsubsection{Results}
\begin{figure}[t]
    \centering
    \includegraphics[width=0.99\columnwidth]{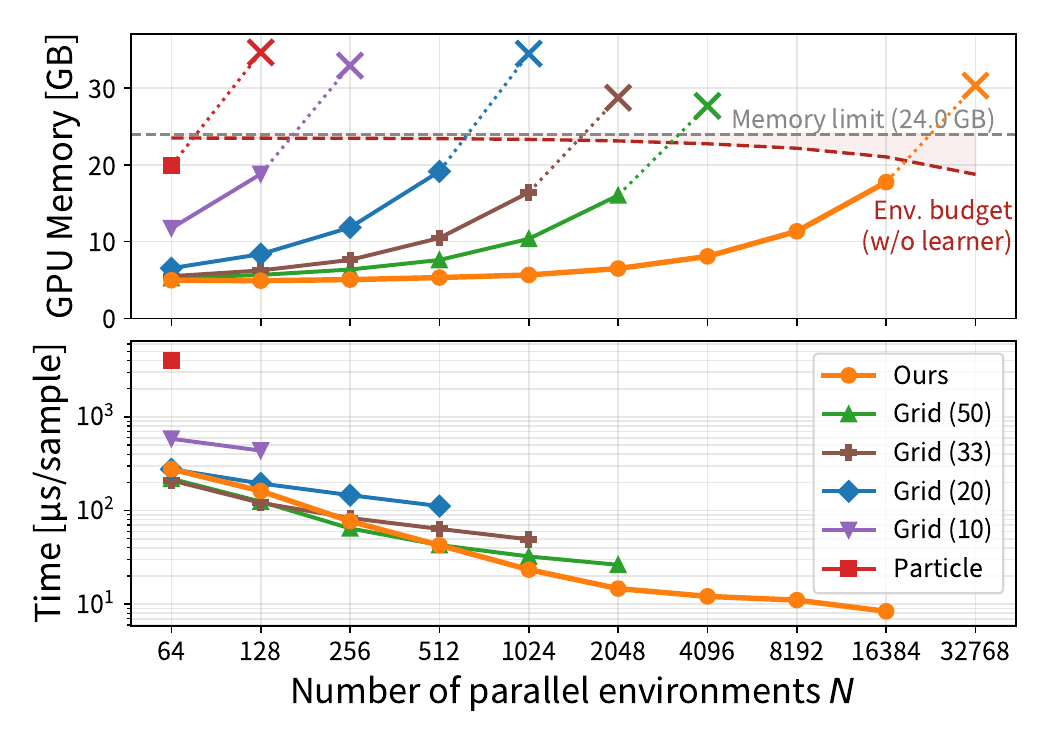}
    \caption{GPU memory usage (top) and simulation time per sample (bottom) versus the number of parallel environments $N$ (logarithmic axis) for Ours, Grid with cell sizes of 50, 33, 20, and 10\,cm (Grid~(50)--Grid~(10)), and Particle; the bottom legend applies to both panels. All representations are measured in the policy-learning environment, with $N$ doubled from 64 until a representation fails to run; each point averages 10 training iterations. (Top) GPU memory in use, excluding the policy learner. Crosses mark the smallest tested $N$ at which a representation failed (out of GPU memory or a stalled simulation), placed at the memory extrapolated from its measured points; the dashed horizontal line marks the 24-GB GPU memory limit, and the red dashed curve shows the memory left for the environment after subtracting the measured PPO-learner memory at the same $N$. (Bottom) Wall-clock simulation time per 50-Hz control step divided by $N$ (logarithmic axis), excluding policy inference and updates.}
    \label{fig:memory_time}
\end{figure}

As shown in \figref{fig:memory_time}, Ours uses the least GPU memory at every $N$ and is the only representation that runs up to $N=16{,}384$. At small $N$, Ours is not faster per sample than the coarse Grids, but it becomes faster than them as $N$ grows; at the largest $N$ of each representation, its time per sample is about one-third of that of Grid (50) and less than 1/400 of that of Particle. These results indicate that the advantage of Ours in time per sample arises as $N$ grows and is largest at the $N$ values that only Ours can run in this measurement.

\subsection{RQ2: Training Time to an Effective Retreat Policy}\label{subsec:exp-learning-efficiency}

\subsubsection{Settings}
To examine whether hybrid terrain modeling shortens the training time to an effective retreat policy compared with individual rigid terrain, we train Ours and Grid with cell sizes from coarse to fine and evaluate the training time and the policy effectiveness by comparing their times to convergence and sim-to-sim success rates, respectively. Although a larger number of parallel environments $N$ shortens the simulation time per sample (\figref{fig:memory_time}), it also collects more samples per rollout and thus updates the policy less frequently per sample, so it does not necessarily improve policy performance or shorten convergence. We therefore also compare the times to convergence of Ours with different $N$. The $N$ used for Ours in the comparison with Grid is also used in all other experiments.

\subsubsection{Results}
\begin{figure}[t]
    \centering
    \includegraphics[width=0.95\columnwidth]{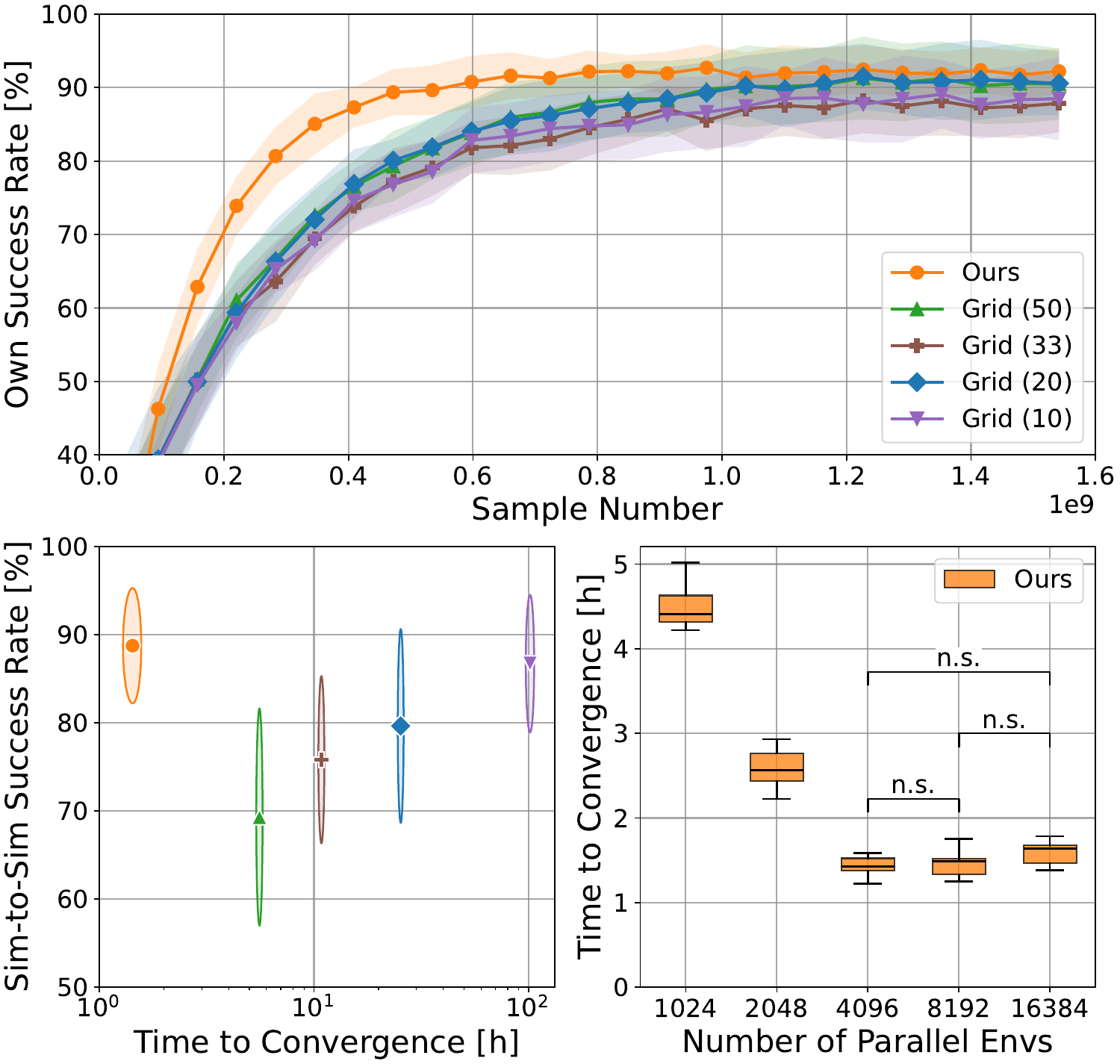}
    \caption{Learning curves (top) and time to convergence (bottom) of Ours and Grid~(50)--Grid~(10). Time to convergence is the number of samples at which a run first reaches $y_1 + 0.95(\bar{y}^{\infty} - y_1)$, where $y_1$ is its first point and $\bar{y}^{\infty}$ is the mean of its last five points (linearly interpolated between points), multiplied by the simulation time per sample in \figref{fig:memory_time} (bottom). (Top) Own success rate, i.e., the success rate logged during training in each method's own training environment, versus the number of samples. Ours uses 4,096 parallel environments, and Grid~(50)--Grid~(10) use 2,048, 1,024, 512, and 128, the largest $N$ at which each ran in \figref{fig:memory_time}, accumulating rollouts until $4{,}096\times24$ samples are collected before each policy update, as Ours does. Lines and shaded bands show the mean and standard deviation over five runs. (Bottom left) Sim-to-sim success rate, i.e., the Average column of \figref{fig:sim2sim_success_rate}, versus time to convergence (logarithmic axis); markers follow the top legend, and ellipses show the standard deviation along each axis. (Bottom right) Time to convergence of Ours with 1,024--16,384 parallel environments (five runs each), updating the policy after every rollout; boxes show the median and interquartile range, whiskers reach the most extreme run within 1.5 interquartile ranges of the box, and brackets show pairwise two-sided Welch's $t$-tests among 4,096, 8,192, and 16,384 environments (n.s.: $p \geq 0.05$).}
    \label{fig:plot_learning_time_vs_resolution}
\end{figure}

As shown in \figref{fig:plot_learning_time_vs_resolution}, all methods reach similar final own success rates, but Ours converges with fewer samples. Combined with its shorter time per sample (\figref{fig:memory_time}), this reduces the time to convergence of Ours to about one-fourth of that of Grid (50) and about 1/70 of that of Grid (10). The sim-to-sim success rate of the Grid baselines increases with finer cells only at the cost of longer convergence, whereas Ours reaches about 90\%, comparable to that of the finest Grid, with the shortest time to convergence. In the sweep over $N$, the time to convergence of Ours decreases up to 4,096 environments, and Welch's $t$-tests detect no significant difference for $N\geq4{,}096$; we therefore select $N=4{,}096$.

\subsection{RQ3: Effect of Each Collapse Primitive on Policy Performance}\label{subsec:exp-collapse-ablation}

\subsubsection{Settings}
To evaluate how each collapse primitive used in training contributes to retreat performance, we compare the retreat success rates of Ours, trained with all three collapse primitives, and of three ablations, each trained without one of them, under each of the three primitives. Ours is thus tested only under primitives used in its training, whereas each ablation is also tested under the omitted primitive, which is not used in its training. Retreat under collapse represented differently from that in training, by rigid blocks or particles in simulation or by physical materials on the real robot, is evaluated in \chapref{subsec:exp-sim2sim} and \chapref{subsec:exp-real-transfer}, respectively.

\subsubsection{Results}
\begin{figure}[t]
    \centering
    \includegraphics[width=0.99\columnwidth]{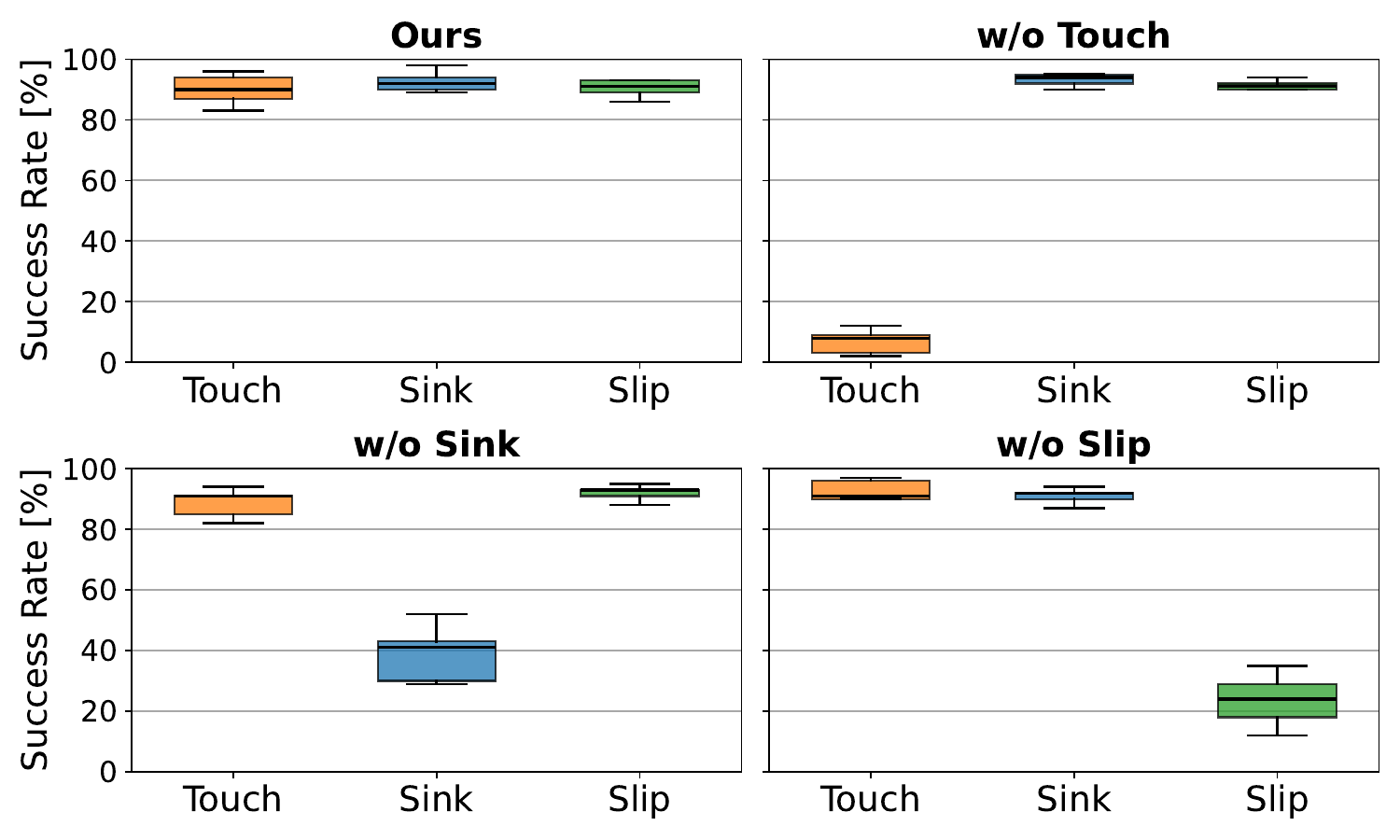}
    \caption{Retreat success rates in the collapse-primitive ablation: Panels show the training conditions, namely Ours and the ablations w/o Touch, w/o Sink, and w/o Slip, which otherwise use the same training settings as Ours. In each panel, the horizontal axis indicates the collapse primitive used for testing; for each training condition, five independently trained policies are each tested with 1,000 trials under each primitive, and each box summarizes their five success rates, with boxes and whiskers drawn as in \figref{fig:plot_learning_time_vs_resolution} (bottom right).}
    \label{fig:plot_ablation_success_rate_heatmap}
\end{figure}

As shown in \figref{fig:plot_ablation_success_rate_heatmap}, Ours achieves median success rates of about 90\% under all three tested collapse primitives. In contrast, each ablation fails mainly under the primitive omitted from its training, where its median success rate drops to about 40\% or lower; the drop is largest for w/o Touch under Touch, where the median success rate is below 10\%. Under the two primitives included in their training, however, the ablations achieve success rates comparable to those of Ours. This suggests that, although all three primitives end in support loss, retreat learned under two of them transfers only partially to the omitted one, i.e., that each primitive provides contact responses that the other two do not cover.

\subsection{RQ3: Retreat Performance Across Terrain Types}\label{subsec:exp-complex-terrain}

\subsubsection{Settings}
To analyze how the retreat performance of the learned policy differs across terrain types, we test Ours on each of the four terrain types used for training (\figref{fig:terrain_curriculum}). We also analyze the causes of failure by classifying each failed trial as Fall, in which the robot is caught in the collapse and falls, or False Retreat, in which the robot mistakes the surrounding rigid terrain for a collapse and retreats from it.

\subsubsection{Results}
\begin{figure}[t]
    \centering
    \includegraphics[width=0.95\columnwidth]{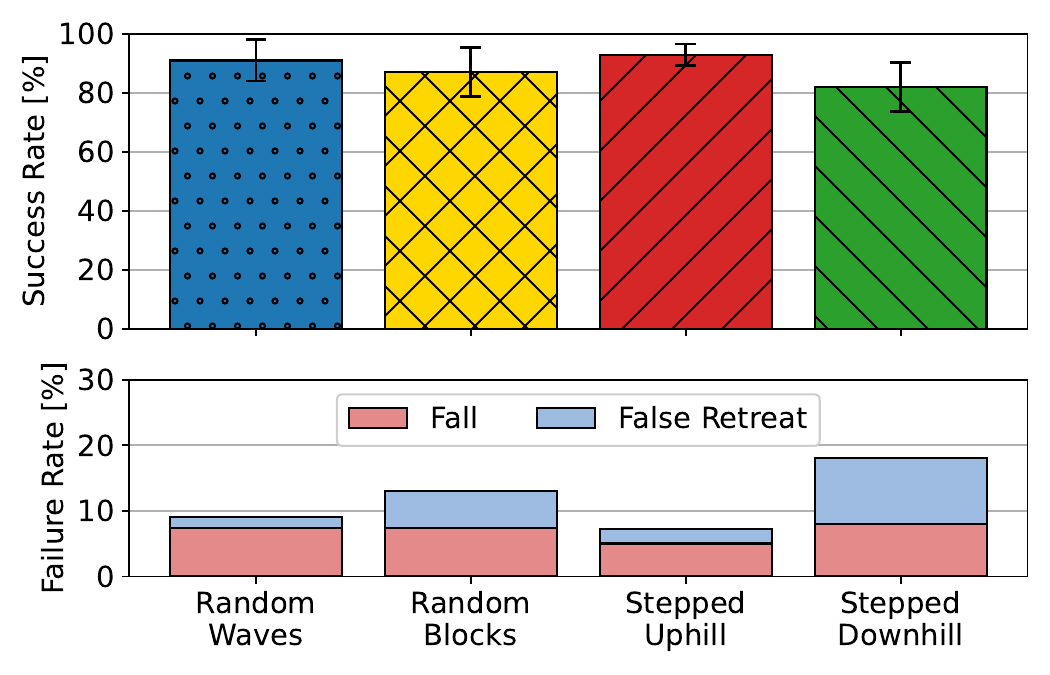}
    \caption{Retreat success rate (top) and failure breakdown (bottom) of Ours across the four training terrain types (\figref{fig:terrain_curriculum}). For each terrain type, five independently trained policies are each tested with 1,000 trials, and each trial is counted as exactly one of Success, Fall, or False Retreat. (Top) Bars show the mean success rate over the five policies, and error bars indicate the standard deviation. (Bottom) Stacked bars show the mean Fall and False Retreat rates; each stack equals 100\% minus the mean success rate.}
    \label{fig:complex_sim}
\end{figure}

As shown in \figref{fig:complex_sim}, Ours achieves mean success rates of over 80\% on all four terrain types, and its mean False Retreat rate is at most 10\% on every terrain type. This indicates that the policy retreats from collapse on all four terrain types while mostly traversing the surrounding terrain without retreat. However, Stepped Downhill has the lowest success rate, and the failure breakdown in \figref{fig:complex_sim} (bottom) shows that it is the only terrain type on which False Retreat accounts for more failures than Fall, with a higher False Retreat rate than on the other terrain types. Because Stepped Downhill is used in training, these false retreats are unlikely to stem from a lack of training on this terrain type; rather, this suggests that, without exteroceptive sensing, the descent onto a traversable downward step can produce proprioceptive responses similar to those of a collapse.

\begin{figure*}[t]
    \centering
    \includegraphics[width=\textwidth]{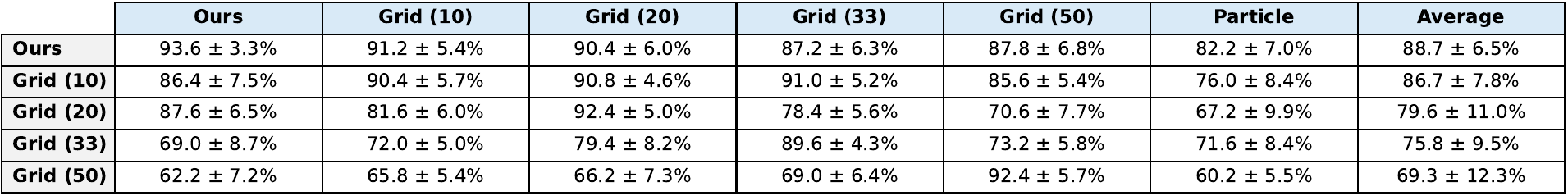}
    \caption{Sim-to-sim retreat success rates across terrain representations: Rows indicate the representation used for training, and columns indicate the representation used for testing; Particle is used only for testing. Each entry in the six testing columns is the mean $\pm$ standard deviation over the five independently trained policies of \chapref{subsec:exp-learning-efficiency}, each tested with 100 trials. The Average column pools the 30 success rates of each row (six testing representations $\times$ five policies), including the testing representation that matches the training one, and reports their mean $\pm$ standard deviation.}
    \label{fig:sim2sim_success_rate}
\end{figure*}

\begin{figure}[t]
    \centering
    \includegraphics[width=0.99\columnwidth]{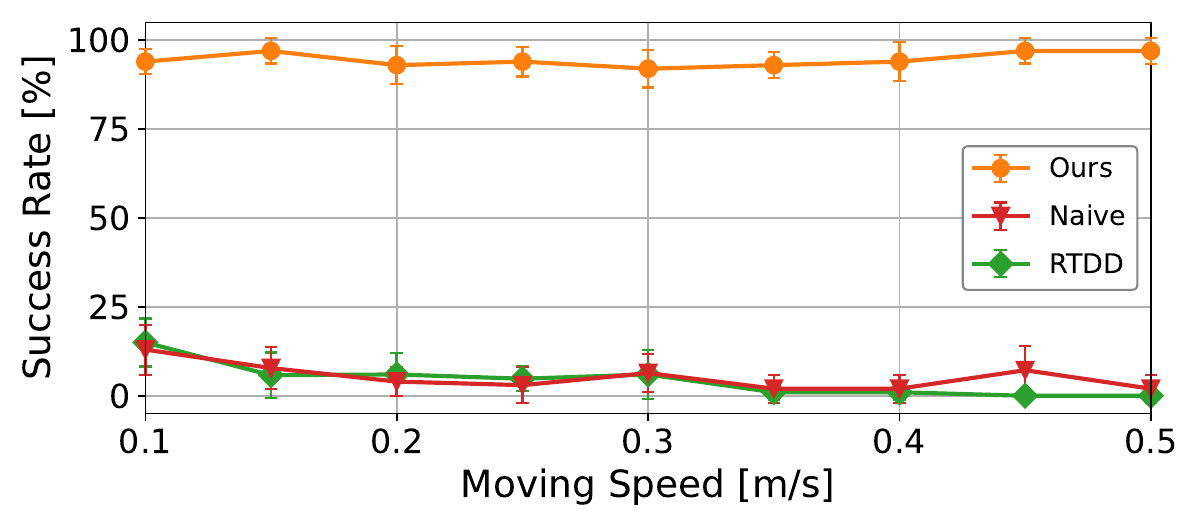}
    \caption{Moving speed versus success rate under sudden ground collapse. The horizontal axis is the measured moving speed, i.e., the mean forward base speed from 1\,s after the approach starts until a wheel enters the collapse-capable opening, and the success rate is the percentage of trials without a fall within 3\,s after the collapse starts. All methods are evaluated in the same scenario with domain randomization and observation noise: the robot drives straight into a $2\,\mathrm{m}\times2\,\mathrm{m}$ collapse-capable opening whose randomly shaped collapse region loses support 0.05--0.20\,s after a wheel touches it (Touch); the forward command is kept for 1--3\,s after the collapse starts and is followed by a retreat command, which Naive and RTDD issue earlier when their estimators detect the collapse. Trials with swept forward commands are binned within $\pm$0.01\,m/s of 0.10, 0.15, \dots, 0.50\,m/s. For each method, five independently trained policies are each evaluated with 100 trials per bin; points and error bars show the mean and standard deviation across the five policies.}
    \label{fig:plot_speed_vs_retreat}
\end{figure}

\begin{figure}[t]
    \centering
    \includegraphics[width=0.5\textwidth]{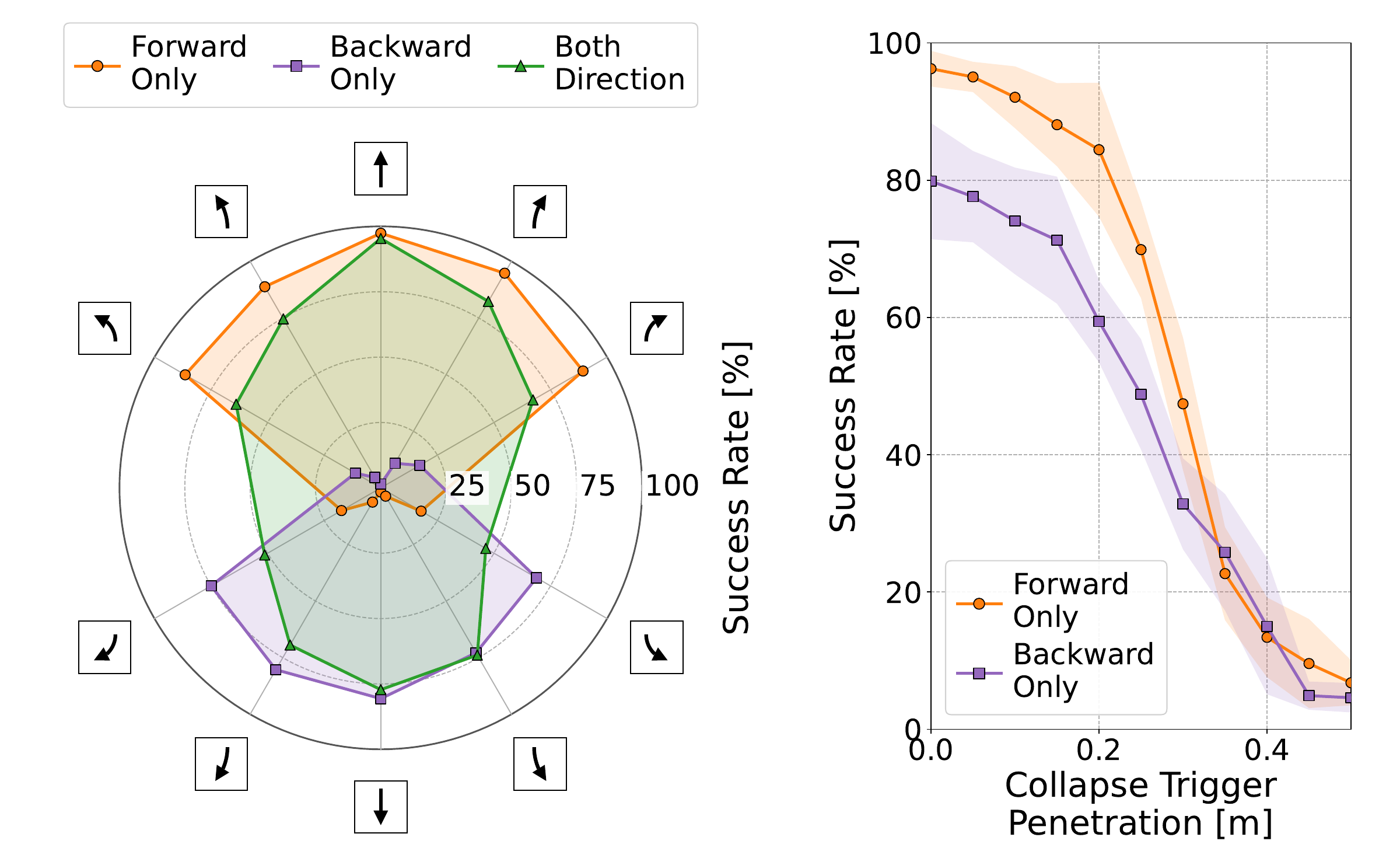}
    \caption{Retreat success versus entry command (left) and collapse trigger penetration (right) on flat ground. (Left) Forward Only, Backward Only, and Both Direction policies, each evaluated after its training converged. The robot approaches the collapse-capable opening straight, switches 0.3--0.5\,m before its edge to one of ten entry commands, holds it after a wheel enters the opening for a random duration as in training (\tabref{tab:common_params}), and then receives a reverse retreat command toward a point 2\,m behind the entry point. Each spoke is an entry command combining a forward (upper half) or backward (lower half) velocity command with a yaw-rate command of $-1$, $-0.5$, 0, 0.5, or 1\,rad/s, sketched by the framed arrow (robot facing upward). Each spoke shows the mean success rate over five independently trained policies, each tested with 100 trials per entry command. (Right) Success rate of Forward Only with straight forward entry and Backward Only with straight backward entry versus the collapse trigger penetration, i.e., the distance the robot advances after a wheel first touches the opening until the collapse is triggered, which varies randomly in training but is fixed to each value in this evaluation, with 100 trials per value; lines and shaded bands show the mean and standard deviation across the five policies.}
    \label{fig:policy_limitation}
\end{figure}

\subsection{RQ4: Sim-to-Sim Transfer Across Terrain Representations}\label{subsec:exp-sim2sim}

\subsubsection{Settings}
To evaluate the generalization of the learned policies to other terrain representations, we test the policies trained with Ours and Grid (50)--Grid (10) in \chapref{subsec:exp-learning-efficiency} without additional training in all six representations, including Particle. Each policy is thus tested in its own training representation and in five representations not used in its training; no policy is trained with Particle.

\subsubsection{Results}
As shown in \figref{fig:sim2sim_success_rate}, Ours and Grid (50)--Grid (10) each achieve mean success rates of about 90\% in the testing representation that matches their training one. However, the Average of the Grid policies decreases as the cell size increases, to about 70\% for Grid (50), whereas Ours achieves the highest Average, about 90\%. Particle is the most difficult testing representation, with the lowest success rate averaged over the training representations, but Ours still achieves a mean success rate of over 80\% in Particle, the highest among the training representations. This suggests that training with a coarser grid discretization limits transfer to terrain representations not used in training.

\begin{figure*}[t]
    \centering
    \includegraphics[width=\textwidth]{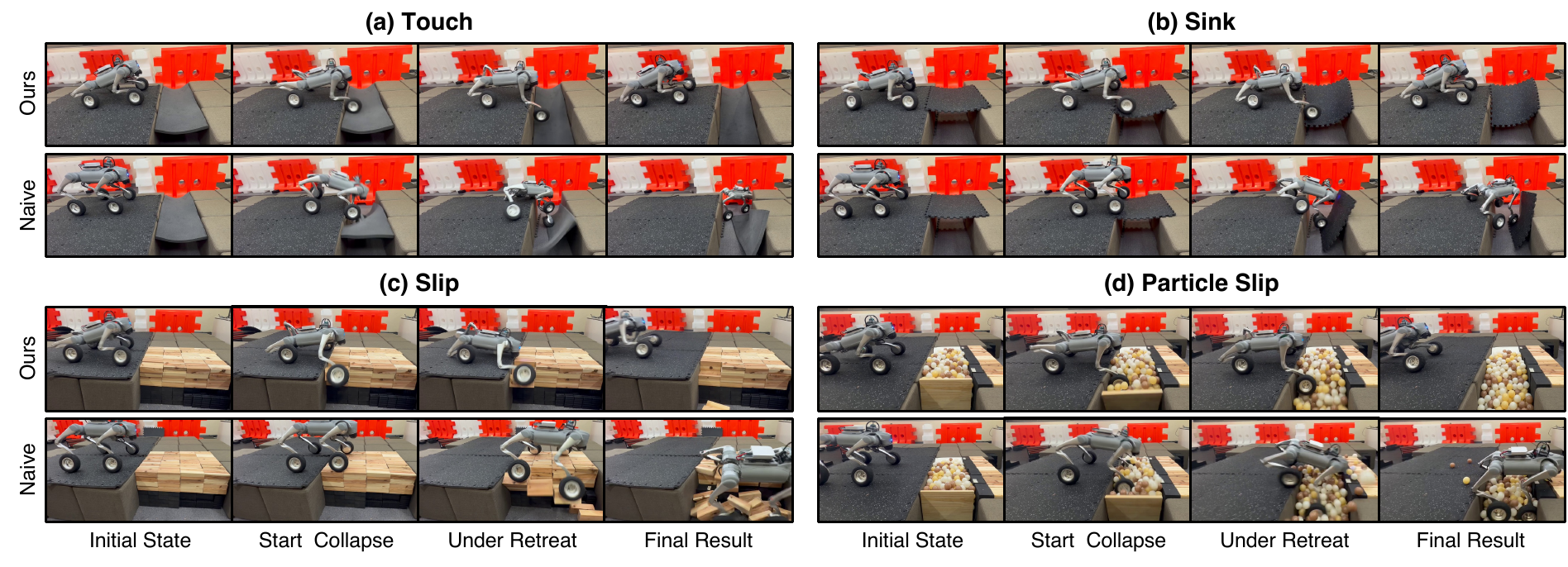}
    \caption{Snapshots of Ours and Naive in the sim-to-real experiment: Panels show the four physical collapse scenarios, (a) Touch, (b) Sink, (c) Slip, and (d) Particle Slip. The test floor is built of tiled blocks with a gap one block wide, and both the blocks and the gap are 40\,cm wide. In (a), the gap is covered with a thin sponge sheet that gives way on contact; in (b), it is covered with a rubber sheet that sinks gradually under the robot and finally gives way; in (c), it is filled with wooden blocks, and the robot drives over their edge so that the blocks slip and fall into the gap; and in (d), it is filled with plastic balls held by a support plate that comes off under load, so that the balls avalanche down. In each panel, the upper row shows Ours and the lower row shows Naive (\chapref{subsubsec:method-baselines}), and the columns show, from left to right, the initial state, the start of the collapse, the robot under retreat, and the final result. Each row shows one representative trial.}
    \label{fig:real_snapshot}
\end{figure*}

\begin{figure}[t]
    \centering
    \includegraphics[width=0.95\columnwidth]{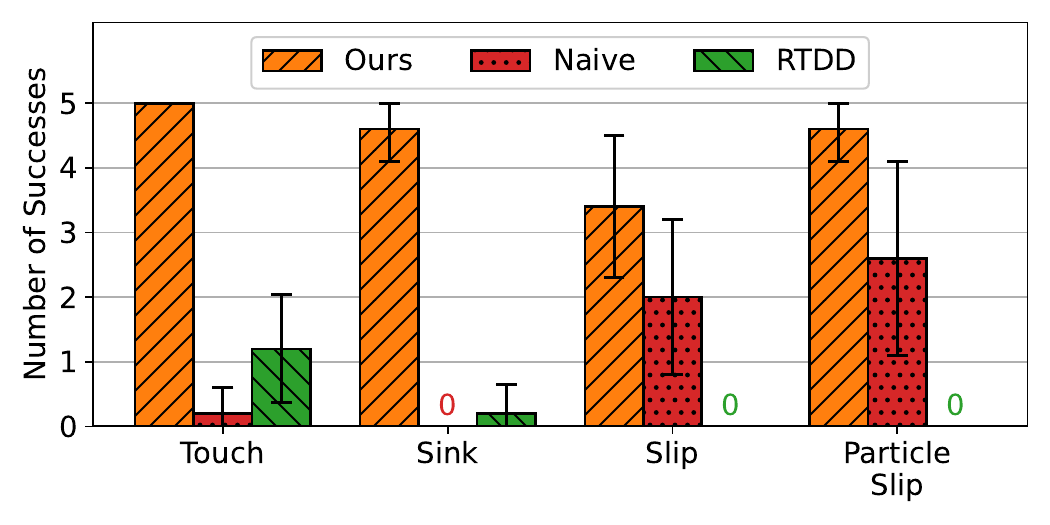}
    \caption{Number of successes of Ours, Naive, and RTDD in the sim-to-real experiment for the four physical collapse scenarios (\figref{fig:real_snapshot}). For each method and scenario, five independently trained policies are each tested five times on the real robot (25 trials per method and scenario). Bars show the mean number of successes per policy out of five trials, and error bars show the standard deviation across the five policies, clipped to 0--5. A ``0'' label indicates zero successes.}
    \label{fig:sim2real-success-rate}
\end{figure}

\subsection{RQ4: Comparison with Estimator-Triggered Retreat Across Traversal Speeds}\label{subsec:exp-speed-escape-balance}

\subsubsection{Settings}
To examine whether the collapse-reactive retreat of Ours outperforms the estimator-triggered retreat of Naive and RTDD, we compare the fall avoidance of the three methods in simulation across traversal speeds.

\subsubsection{Results}
As shown in \figref{fig:plot_speed_vs_retreat}, Ours avoids falls in over 90\% of the trials in every speed bin, whereas Naive and RTDD avoid falls in less than 20\% of the trials in every speed bin. This indicates that Ours maintains a high fall-avoidance rate over the tested speed range, in which the estimator-triggered retreat of Naive and RTDD rarely prevents a fall. Because the estimators of Naive and RTDD can only advance the retreat command, these baselines receive it no later than Ours. Their low fall-avoidance rates therefore suggest that their locomotion policy, which is trained on non-collapsible terrain (\chapref{subsubsec:method-baselines}), cannot keep the robot balanced after support loss, whereas Ours is trained to do so during the hold phase (\chapref{subsubsec:episode-commands}).

\subsection{RQ5: Effect of Collapse-Exposure Design}\label{subsec:exp-policy-limitations}

\subsubsection{Settings}
To examine the effect of the collapse-exposure design, we compare policies trained with collapse encountered only during forward motion (Forward Only, trained with the same settings as Ours), only during backward motion (Backward Only), and during both forward and backward motion (Both Direction). We evaluate the retreat success rate of each policy for entry commands in each direction of motion, and the dependence on the timing of the collapse by comparing the retreat success rates of the direction-specific policies in their trained directions under different collapse trigger penetrations.

\subsubsection{Results}
As shown in \figref{fig:policy_limitation} (left), retreat follows the direction of collapse exposure in training: Forward Only succeeds in over 85\% of the trials for every forward entry command but in less than 20\% for every backward one, and Backward Only shows the opposite tendency with a lower mean success rate after entry in its trained direction (about 75\%, versus about 90\% for Forward Only). Both Direction retreats after either entry, but its mean success rate in each direction is more than 10 percentage points lower than that of the direction-specific policy, mostly after entry while turning.

As shown in \figref{fig:policy_limitation} (right), success also depends on the collapse trigger penetration, i.e., the distance the robot advances from first touching the opening until the collapse: Forward Only stays above 80\% up to 0.2\,m, whereas Backward Only already decreases within this range, and both rarely succeed at the largest penetrations, likely because less stable support then remains for reversing motion. These results indicate a trade-off in the collapse-exposure design: forward-only exposure, used in Ours, gives the highest success after forward entry and thus suits a robot that mainly advances into unknown terrain, whereas exposure in both directions enables retreat after either entry at the cost of lower success.

\subsection{RQ6: Sim-to-Real Transfer Across Four Physical Collapse Scenarios}\label{subsec:exp-real-transfer}

\subsubsection{Settings}
To evaluate whether the retreat behavior learned in simulation transfers to physical collapse scenarios, we deploy the simulation-trained policies of Ours, Naive, and RTDD on the real Go2W robot without additional training. We compare their numbers of successful retreats across the four experimentally constructed collapse scenarios shown in \figref{fig:real_snapshot}.

\subsubsection{Results}
As shown in \figref{fig:sim2real-success-rate}, Ours achieves the largest mean number of successes in all four scenarios, with on average more than four successes out of five trials under Touch, Sink, and Particle Slip and the fewest under Slip. Naive succeeds in almost none of the trials under Touch and Sink and in about half of the trials or fewer under Slip and Particle Slip, and RTDD fails in most trials in all four scenarios. \figref{fig:real_snapshot} shows representative trials in which Ours retreats onto the non-collapsible floor, whereas Naive falls into the collapsed region. This indicates that the retreat behavior learned in simulation transfers to the physical collapse scenarios on the real robot without additional training, whereas the estimator-triggered external retreat commands of Naive and RTDD are insufficient for reliable retreat, particularly under Touch and Sink.

\subsection{Outdoor Demonstration}\label{subsec:exp-outdoor}
\begin{figure}[t]
    \centering
    \includegraphics[width=0.99\columnwidth]{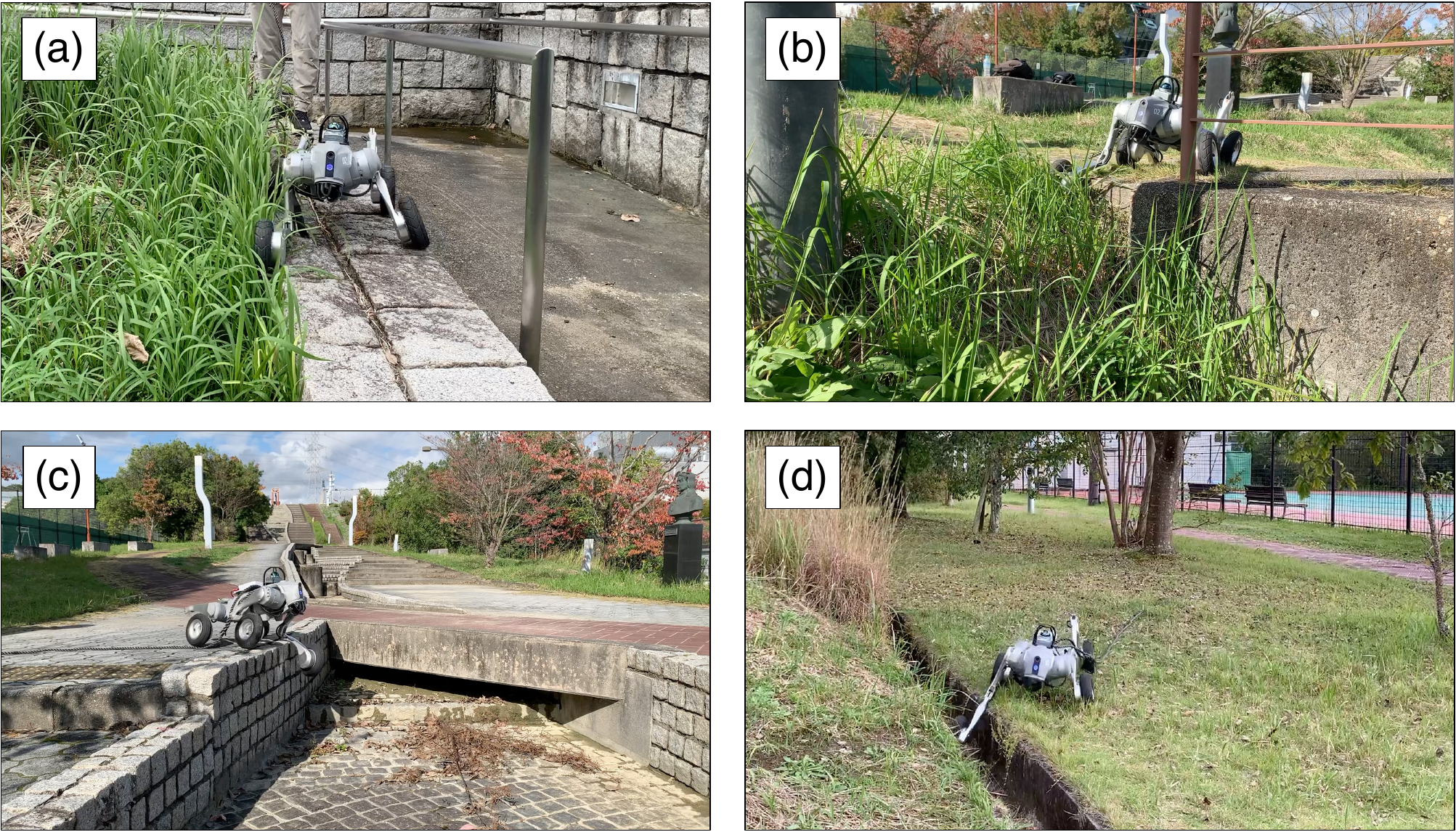}
    \caption{Snapshots of the outdoor demonstration of Ours on the real Go2W robot: (a) the edge of a path bordered by grass, (b) the grass-covered edge of a concrete wall, (c) the edge of a stone wall above a channel, and (d) a ditch in a lawn.}
    \label{fig:outdoor}
\end{figure}

In addition to the quantitative evaluations, we qualitatively demonstrate the policy of Ours on the real Go2W robot in outdoor environments without additional training. The robot was operated at four sites where a wheel can suddenly lose support, such as the edges of paths and walls and a ditch (\figref{fig:outdoor}). At these sites, the robot retreated to the supported ground after a wheel lost support. Details are shown in the supplementary video.

\section{Discussion}

\subsection{Scalability and Extensibility of the Proposed Framework}
The computational benefits of hybrid terrain modeling grow with the number of parallel environments (\figref{fig:memory_time}). However, increasing this number beyond 4,096 did not significantly shorten the time to convergence (\figref{fig:plot_learning_time_vs_resolution}), suggesting the importance of balancing simulation throughput and policy-update frequency for efficient learning. Beyond computational efficiency, the framework may also apply to other robot morphologies, because the SMCM formulation (\chapref{subsubsec:shared-contact-law}) is not inherently limited to wheeled quadrupeds. The current implementation applies virtual contact forces only to the wheels, which sufficed for the collapse-reactive retreat task considered here (\figref{fig:sim2sim_success_rate} and \figref{fig:sim2real-success-rate}). For tasks that require whole-body interaction, the SMCM could be extended to other robot links by computing their contact positions, penetration depths, and relative velocities with respect to the virtual ground, still without duplicating terrain objects across parallel environments.

\subsection{Particle-Based Simulation and Fidelity of Virtual Collapse Modeling}
Particle-based simulation captures material motion during ground collapse but is costly: Particle ran only at the smallest tested $N$ (\figref{fig:memory_time}) because its densely packed rigid-body particles incur high memory and contact-computation costs. Faster granular solvers~\cite{millard2023granulargym}, coarse-to-fine resolution transfer~\cite{kadokawa2025prpd}, and particle counts scaled with task difficulty or restricted to collapsing regions~\cite{kadokawa2026obstacle} can lower this cost, but particles must still be simulated throughout training. Our sim-to-sim results suggest that this may be unnecessary: a policy trained only with hybrid terrain modeling achieved over 80\% success in Particle (\figref{fig:sim2sim_success_rate}). However, this transfer does not establish the physical fidelity of the virtual contact model, whose correspondence to real ground responses remains unquantified despite the randomized contact and collapse parameters (\tabref{tab:dr}). Future comparisons of contact forces, support-loss timing, and robot motion between simulated and real ground failure could reveal which aspects matter most for policy learning, and thus where to refine the virtual contact model and where particle simulation is worth its cost.

\subsection{Role of Collapse-Response Diversity in Policy Learning}
The collapse-primitive ablation study (\chapref{subsec:exp-collapse-ablation}, \figref{fig:plot_ablation_success_rate_heatmap}) compared training with Touch, Sink, and Slip against training with each primitive omitted. Removing one primarily reduced retreat success under that primitive, while success under the other two remained comparable to Ours. This suggests limited generalization across contact-response patterns, rather than an overall inability to learn retreat. Excluding Touch caused the largest degradation, possibly because support loss without preceding subsidence or a lateral push requires reactions not acquired from Sink or Slip. These results highlight the importance of training with qualitatively different collapse responses rather than relying on a single primitive. The SMCM supports this diversity by allowing additional collapse primitives to be modeled through state variables and transitions without introducing terrain objects. However, whether the three primitives considered here are sufficient to generalize to other types of ground failure remains an open question.

\subsection{Proprioceptive Sensing: Strengths and Limitations}
Relying only on proprioception, the policy responds to the robot's reaction to terrain failure rather than to the terrain state, enabling collapse-reactive retreat without prior terrain inspection or a separate collapse detector, which may help when visual information is unreliable. However, this reliance on indirect cues makes ground failure hard to distinguish from traversable terrain changes with similar proprioceptive responses. In the terrain-type evaluation (\chapref{subsec:exp-complex-terrain}, \figref{fig:complex_sim}), the mean false-retreat rate was at most 10\% on every terrain type, but Stepped Downhill had the highest rate and was the only terrain on which false retreats outnumbered falls, possibly because descending a step produces foot and body motion similar to that caused by ground collapse. Exteroceptive terrain assessment or tactile probing~\cite{haddeler2022traversability,miki2022robust,SPS,wellhausen2019where,kolvenbach2019haptic,fankhauser2018probabilistic} could complement proprioception by distinguishing such changes from ground failure, reducing false retreats while retaining the response to collapse that becomes apparent only after contact.

\subsection{Collapse-Reactive Retreat on High Steps}
The terrain-type evaluation (\chapref{subsec:exp-complex-terrain}, \figref{fig:complex_sim}) showed mean retreat success rates above 80\% across all four terrain types, but the stepped terrains had risers of at most 0.08\,m (\figref{fig:terrain_curriculum}). This leaves open whether the learned policy can retreat after collapse on terrain with larger height discontinuities. High steps introduce a challenge beyond ordinary terrain traversal: the robot may need to lift or reposition its legs to negotiate a step while simultaneously compensating for the loss of ground support. In a wheeled quadruped, these demands may compete for the same leg motions and available contacts, particularly when retreating across the steps just traversed. Future work should extend the terrain curriculum to higher steps and evaluate normal traversal and post-collapse retreat separately. Such experiments would help distinguish limitations in step-negotiation ability from those in collapse-reactive retreat and clarify whether the two behaviors can be learned jointly.

\subsection{Collapse-Reactive Retreat versus Estimator-Triggered Retreat}
The comparison with estimator-triggered retreat (\chapref{subsec:exp-speed-escape-balance}, \figref{fig:plot_speed_vs_retreat}) suggests that detecting collapse and issuing a retreat command may be insufficient if the underlying locomotion policy cannot handle support loss. Naive and RTDD avoided falls in fewer than 20\% of trials across the tested speeds, despite issuing retreat commands upon collapse detection. In contrast, Ours exceeded 90\% even when the forward command was maintained briefly after collapse. This suggests that maintaining balance during support loss may be more critical than retreat-command timing alone. By learning locomotion and collapse-reactive retreat within a single policy, Ours can acquire control responses to changing contact conditions. The real-robot results (\chapref{subsec:exp-real-transfer}, \figref{fig:sim2real-success-rate}) further support this interpretation, with Ours achieving the highest success across all four physical collapse scenarios.

\section{Conclusion}
This paper proposed a memory-efficient parallel policy-learning framework that trains a single proprioceptive policy for normal locomotion and collapse-reactive retreat via hybrid terrain modeling. Its SMCM computes virtual contact forces from a shared contact-response law and independent collapse states, preventing indirect interference without creating terrain objects for the collapsible region in each environment. Compared with fine-resolution rigid-block terrain, the framework supported over 100 times as many parallel environments within the same GPU memory limit and achieved over 50 times faster policy convergence with comparable retreat performance. The policy transferred without additional training to particle-based simulation and four physical collapse scenarios and was also demonstrated outdoors, suggesting that task-relevant contact responses can provide the state--action experiences needed to learn collapse-reactive retreat without detailed ground-failure simulation.

\appendices
\begin{table}[t]
    \centering
    \caption{Observations of the Actor and the Critic}
    \label{tab:observations}
    \footnotesize
    \setlength{\tabcolsep}{3pt}
    \begin{tabularx}{\columnwidth}{@{}Yccc@{}}
        \toprule
        Observation & Dim. & Actor & Critic \\
        \midrule
        Base angular velocity in the body frame & 3 & \cmark & \cmark \\
        Projected gravity in the body frame & 3 & \cmark & \cmark \\
        Velocity command (forward, lateral, and yaw rate) & 3 & \cmark & \cmark \\
        Joint positions relative to the default posture (wheel entries set to zero) & 16 & \cmark & \cmark \\
        Joint velocities & 16 & \cmark & \cmark \\
        Previous action & 16 & \cmark & \cmark \\
        Base linear velocity in the body frame & 3 & \xmark & \cmark \\
        Signed distance from the base to the opening boundary (positive outside) & 1 & \xmark & \cmark \\
        Signed distance from each wheel to the opening boundary (positive inside) & 4 & \xmark & \cmark \\
        Base height above the local terrain surface & 1 & \xmark & \cmark \\
        Height of each wheel above the local terrain surface & 4 & \xmark & \cmark \\
        Contact state of each wheel & 4 & \xmark & \cmark \\
        \midrule
        Total dimensions & & 57 & 74 \\
        \bottomrule
    \end{tabularx}
\end{table}

\begin{table}[t]
    \centering
    \caption{Common Parameters Used to Train Ours}
    \label{tab:common_params}
    \footnotesize
    \begin{tabularx}{\columnwidth}{@{}>{\raggedright\arraybackslash}p{0.42\columnwidth}Y@{}}
        \toprule
        Parameter & Value \\
        \midrule
        \multicolumn{2}{@{}l}{\textit{Episode and commands}} \\
        Initial distance & $\mathcal{U}[3.0,\,3.4]$\,m from the opening center \\
        Initial approach angle & $\mathcal{U}[-25^{\circ},\,25^{\circ}]$ from the approach axis \\
        Initial heading error & $\mathcal{U}[-0.15,\,0.15]$\,rad \\
        Approach and retreat speeds & $\mathcal{U}[0.1,\,0.5]$\,m/s \\
        Heading-hold yaw-rate gain & 1.5\,s$^{-1}$ on the heading error \\
        Heading-hold yaw-rate limit & $[-1,\,1]$\,rad/s \\
        Hold-phase duration & $\mathcal{U}[1,\,3]$\,s \\
        Hold-phase stop command & With probability 0.4 \\
        Initial stop command & First 2\,s of each episode \\
        Inserted stop duration & $\mathcal{U}[1,\,2]$\,s, once per episode \\
        Inserted stop onset & $\mathcal{U}[1,\,6]$\,s after the episode start \\
        \midrule
        \multicolumn{2}{@{}l}{\textit{Terrain and curriculum}} \\
        Terrain-tile mesh & $48\times48$ vertices \\
        Rigid-terrain friction & 1.0 \\
        Rigid-terrain restitution & 0 \\
        Initial terrain level & $\ell=1$ \\
        Demotion without encounter & After four consecutive episodes \\
        Uniform level sampling & 35\% of the resetting environments \\
        \midrule
        \multicolumn{2}{@{}l}{\textit{Collapse primitives}} \\
        Effective mass $m_i$ & One quarter of the robot mass \\
        Force limit $f_{\max}$ & 500\,N \\
        Slip-velocity scale $\varepsilon$ & 0.05\,m/s \\
        Subsidence-rate limit $\dot{A}_{\max}$ & 0.30\,m/s \\
        Number of harmonics $K$ & 3 \\
        Boundary width $\lambda$ & 0.02\,m \\
        \bottomrule
    \end{tabularx}
\end{table}

\begin{table}[t]
    \centering
    \caption{Domain Randomization and Observation Noise}
    \label{tab:dr}
    \footnotesize
    \setlength{\tabcolsep}{3pt}
    \begin{tabularx}{\columnwidth}{@{}Yl@{}}
        \toprule
        Parameter & Range \\
        \midrule
        \multicolumn{2}{@{}l}{\textit{Physical properties (once per environment)}} \\
        Static friction of the robot bodies & $\mathcal{U}[0.3,\,1.0]$ \\
        Dynamic friction of the robot bodies & $\mathcal{U}[0.3,\,0.8]$ \\
        Restitution of the robot bodies & $\mathcal{U}[0,\,0.5]$ \\
        Base-mass offset & $\mathcal{U}[-1.0,\,1.5]$\,kg \\
        Mass scale of each other link & $\mathcal{U}[0.7,\,1.3]$ \\
        Base center-of-mass offset (each axis) & $\mathcal{U}[-0.05,\,0.05]$\,m \\
        \midrule
        \multicolumn{2}{@{}l}{\textit{Actuation and disturbances}} \\
        Scale of $K_p$ and $K_d$ of each joint, including the wheel damping gain (every reset) & $\mathcal{U}[0.7,\,1.4]$ \\
        Push: base-velocity change in world $x$ and $y$, every $\mathcal{U}[10,\,15]$\,s & $\mathcal{U}[-0.3,\,0.3]$\,m/s \\
        \midrule
        \multicolumn{2}{@{}l}{\textit{Collapse properties (every episode)}} \\
        Friction $\mu$ of the virtual floor & $\mathcal{U}[0.3,\,1.0]$ \\
        Stiffness $k$ of the virtual floor & $\mathcal{U}[1.0,\,2.0]\times10^{4}$\,N/m \\
        Damping ratio $\zeta$ of the virtual floor & $\mathcal{U}[0.5,\,1.0]$ \\
        Trigger force $f_{\mathrm{tr}}$ (support loss and slip) & $\mathcal{U}[5,\,20]$\,N \\
        Collapse delay $\tau$ (support loss) & $\mathcal{U}[0.05,\,0.20]$\,s \\
        Yield force $F_y$ (subsidence) & $\mathcal{U}[3,\,20]$\,N \\
        Subsidence gain $\gamma$ & $\mathcal{U}[0.003,\,0.012]$\,m/(s\,N) \\
        Collapse amplitude $A_c$ (subsidence) & $\mathcal{U}[0.08,\,0.16]$\,m \\
        Nominal slip force $F_s$ & $\mathcal{U}[60,\,140]$\,N \\
        Slip-force scale (every simulation step) & $\mathrm{clip}(1+\mathcal{N}(0,\,0.3^2),\,0,\,2)$ \\
        Slip direction $\mathbf{d}$ (random side; perturbed every simulation step) & $\mathcal{N}(0,\,0.30^2)$\,rad \\
        Slip duration $T$ & $\mathcal{U}[0.3,\,0.8]$\,s \\
        Depth of the slip depression & $\mathcal{U}[0.08,\,0.20]$\,m \\
        Base radius $r_0$ of the collapse region & $\mathcal{U}[0.75,\,1.0]$\,m \\
        Shape coefficients $a_n$, $b_n$ of the collapse region & $\mathcal{N}(0,\,(0.30/n)^2)$ \\
        Center offset of the collapse region (each axis) & $\mathcal{U}[-0.12,\,0.12]$\,m \\
        \midrule
        \multicolumn{2}{@{}l}{\textit{Initial state (every reset)}} \\
        Leg-joint position offset & $\mathcal{U}[-0.2,\,0.2]$\,rad \\
        Leg-joint velocity & $\mathcal{U}[-0.5,\,0.5]$\,rad/s \\
        Base roll and pitch & $\mathcal{U}[-0.1,\,0.1]$\,rad \\
        \midrule
        \multicolumn{2}{@{}l}{\textit{Actor observation noise (every control step)}} \\
        Base angular velocity & $\mathcal{U}[-0.2,\,0.2]$\,rad/s \\
        Projected gravity & $\mathcal{U}[-0.05,\,0.05]$ \\
        Joint positions & $\mathcal{U}[-0.01,\,0.01]$\,rad \\
        Joint velocities & $\mathcal{U}[-1.5,\,1.5]$\,rad/s \\
        \bottomrule
    \end{tabularx}
\end{table}

\begin{table}[!t]
    \centering
    \caption{Reward Terms and Weights}
    \label{tab:reward}
    \scriptsize
    \renewcommand{\arraystretch}{1.1}
    \setlength{\tabcolsep}{3pt}
    \begin{tabularx}{\columnwidth}{@{}>{\raggedright\arraybackslash}p{0.38\columnwidth}Y@{}}
        \toprule
        Term (weight $w_k$) & Definition $r_k$ \\
        \midrule
        \multicolumn{2}{@{}l}{\textit{Command tracking}} \\
        Forward-velocity tracking ($2.0$) & $\exp\!\bigl(-[(\hat v_x-v_x)^2+v_y^2]/0.25\bigr)\,u\,(1-D)$ \\
        Yaw-rate tracking ($2.0$) & $\bigl[\exp\!\bigl(-(\hat\omega_z-\omega_z)^2/0.1\bigr)-|\hat\omega_z-\omega_z|\bigr]$ $u\,(1-D)$ \\
        Forward progress ($8.0$) & $\mathbf{1}[c_x>0.1]\,(1-D)\,\mathrm{clip}(v_x,0,0.5)\,u\,(1+\bar\ell)$ \\
        Lateral drift ($-0.3$) & $v_y^2\,u\,(1-D)$ \\
        \midrule
        \multicolumn{2}{@{}l}{\textit{Collapse response and retreat}} \\
        Hold after collapse ($3.0$) & $D\,(1-B)\,u$ \\
        Retreat progress ($6.0$) & $B\,\mathrm{clip}(-v_x,0,0.5)\,u$ \\
        Wheels out of the opening ($2.0$) & $D\,B\,\frac{1}{4}\sum_{i}(1-\mathrm{in}_i)\,(1+4\sqrt{\bar\ell})\,u$ \\
        Retreat completion ($20.0$) & $(1+\bar\ell)/\Delta t_{\mathrm{c}}$ at the first step of retreat success, and 0 otherwise \\
        \midrule
        \multicolumn{2}{@{}l}{\textit{Stop command}} \\
        Stand still ($2.5$) & $\mathbf{1}[|c_x|<0.1]\,(1-D)\exp(-v_{xy}^2/0.25)\,u$ \\
        Zero forward speed ($-10.0$) & $\mathbf{1}[|c_x|<0.1]\,(1-D)\,|v_x|\,u$ \\
        Stop hold ($3.0$) & $S\exp(-v_{xy}^2/0.25)\,u$ \\
        Stop hold (tight) ($3.0$) & $S\exp(-v_{xy}^2/0.02)\,u$ \\
        Stop brake ($-3.0$) & $S\,v_{xy}\,u$ \\
        Stop wheel brake ($-4.0$) & $S\,\frac{0.086}{4}\sum_{w}|\dot q_w|\,u$ \\
        Stand tall during stop ($4.5$) & $S\,\mathrm{clip}(h_{\mathrm{b}}/0.285,0,1)\,u$ \\
        \midrule
        \multicolumn{2}{@{}l}{\textit{Posture and leg clearance}} \\
        Upright base ($1.0$) & $(1-g_z)^2$ \\
        Base pitch ($2.5$) & $\exp\!\bigl(-(g_x+\sin 20^{\circ})^2/(2\sin^2 15^{\circ})\bigr)\,u$ \\
        Wheels on the ground ($1.5$) & $(1-D)\,\frac{1}{4}\sum_{i}n_i\,u$ \\
        Backward center of mass ($4.0$) & $\mathrm{clip}\bigl((\bar{\mathbf{p}}_{F}-\mathbf{p}_{C})\cdot\hat{\mathbf{h}},\,-0.3,\,0.3\bigr)\,u$ \\
        Hip abduction ($-0.65$) & $(1-D)\,(1-0.5\bar\ell)\sum_{j\in\mathrm{hips}}(q_j-q^0_j)^2$ \\
        Front-elbow clearance ($-0.8$) & $\max(c_x,0)\,(1-0.5\bar\ell)$ $\sum_{l\in\{\mathrm{FR},\mathrm{FL}\}}\mathrm{clip}\bigl((0.12-\eta_l)/0.12,0,1\bigr)$ \\
        Rear-knee clearance ($-1.0$) & $(1-S)\,(1-D)\,(1-0.5\bar\ell)$ $\sum_{l\in\{\mathrm{RR},\mathrm{RL}\}}\mathrm{clip}\bigl((0.12-\eta_l)/0.12,0,1\bigr)$ \\
        Front-leg crossing during stop ($-20.0$) & $S\max(0.12-y_{\mathrm{F}},0)\,u$ \\
        Front-leg crossing ($-8.0$) & $\max(-y_{\mathrm{F}},0)\,u$ \\
        \midrule
        \multicolumn{2}{@{}l}{\textit{Regularization}} \\
        Vertical base velocity ($-2.0$) & $v_z^2\,u$ \\
        Roll and pitch rates ($-0.05$) & $(\omega_x^2+\omega_y^2)\,u\,(1-0.5\bar\ell)$ \\
        Leg-joint torque ($-2.5\times10^{-5}$) & $\sum_j\tau_j^2$ \\
        Leg-joint acceleration ($-2.5\times10^{-7}$) & $\sum_j\ddot q_j^2$ \\
        Wheel-joint acceleration ($-2.5\times10^{-9}$) & $\sum_w\ddot q_w^2$ \\
        Leg-joint power ($-2.0\times10^{-5}$) & $\sum_j|\dot q_j\tau_j|$ \\
        Leg-joint limits ($-5.0$) & $\sum_j\bigl[\max(q_j^{\mathrm{lo}}-q_j,0)+\max(q_j-q_j^{\mathrm{hi}},0)\bigr]$, with soft limits at 90\% of the joint range \\
        Deviation from the default posture ($-0.2$) & $\lVert\mathbf{q}-\mathbf{q}^0\rVert\,u$ \\
        Action rate ($-0.02$) & $\lVert\mathbf{a}_t-\mathbf{a}_{t-1}\rVert^2\,(1-0.5\bar\ell)$ \\
        Non-wheel contacts ($-1.0$) & $u\sum_{b}\mathbf{1}[\lVert\mathbf{F}^{\mathrm{s}}_b\rVert>1\,\mathrm{N}]$ \\
        Wheel contact force ($-1.5\times10^{-4}$) & $\sum_i\max(\lVert\mathbf{F}^{\mathrm{s}}_i\rVert-100\,\mathrm{N},0)$ \\
        Torque saturation ($-0.5$) & $(1-D)\,(1-B)\,(1-0.25\bar\ell)$ $\sum_j\mathrm{clip}\bigl(|\tau^{\mathrm{c}}_j|/\tau_{\max}-0.7,\,0,\,2.5\bigr)$ \\
        \midrule
        \multicolumn{2}{@{}l}{\textit{Termination}} \\
        Fall ($-4.0$) & $1/\Delta t_{\mathrm{c}}$ at a failure termination, and 0 otherwise \\
        \bottomrule
    \end{tabularx}
\end{table}

\section{Observations of the Actor and the Critic}\label{app:observations}
\tabref{tab:observations} lists the observations of the actor and the critic. The base angular velocity and the joint velocities are scaled by 0.25 and 0.05, respectively; during training, the observation noise in \tabref{tab:dr} is added before this scaling, also to the zeroed wheel entries of the joint positions. The critic receives its observations without noise or scaling.

\section{Training Parameters and Domain Randomization}\label{app:parameters}
The simulation and PPO parameters follow those of Legged Gym~\cite{legged_gym}. For the Go2W, the leg joints are controlled with PD gains of $K_p=25$\,N\,m/rad and $K_d=0.5$\,N\,m\,s/rad and the wheel joints with a damping gain of 0.5\,N\,m\,s/rad, the joint torques are limited to $\tau_{\max}=23.5$\,N\,m, and the actions are scaled by 0.125\,rad for the hip joints, 0.25\,rad for the thigh and calf joints, and 5.0\,rad/s for the wheel joints. \tabref{tab:common_params} lists the remaining training parameters, and \tabref{tab:dr} lists the domain randomization and observation noise used for training, where $\mathcal{U}[a,b]$ denotes the uniform distribution on $[a,b]$ and $\mathcal{N}(\mu,\sigma^2)$ the normal distribution with mean $\mu$ and variance $\sigma^2$.

\section{Reward Function and Termination Conditions}\label{app:reward}
\textit{Reward composition:} The reward at control step $t$ is $r_t=\mathrm{clip}\bigl(\Delta t_{\mathrm{c}}\sum_k w_k r_k,\,-50,\,50\bigr)$, where $\Delta t_{\mathrm{c}}=4\Delta t=0.02$\,s is the control period; the terms $r_k$ and weights $w_k$ are listed in \tabref{tab:reward}, and all constants are in SI units.

\textit{Robot state and command:} $\mathbf{v}$, $\boldsymbol{\omega}$, and $\mathbf{g}$ are the base linear velocity, angular velocity, and projected gravity in the body frame ($g_z=-1$ for a level base), and $v_{xy}=\lVert(v_x,v_y)\rVert$. The normalized velocity command $\mathbf{c}=(c_x,c_y,c_z)$ gives the tracking targets $\hat v_x=0.5c_x$\,m/s and $\hat\omega_z=c_z$\,rad/s, which are set to zero when $|c_x|<0.1$ and $|c_z|<0.2$, respectively, and $\hat v_x$ is clipped to be nonpositive while $D=1$. During a stop command, each component of $\mathbf{c}$ is sampled from $\mathcal{U}[-0.1,0.1]$; otherwise $c_y=0$, and $c_y$ is never tracked.

\textit{Gating variables:} The upright factor is $u=\min(\max(-g_z,0),0.7)/0.7$, and $\bar\ell=\ell/\ell_{\max}$ is the normalized terrain level. The collapse latch $D\in\{0,1\}$ is set when the center of a wheel inside the opening falls more than 0.08\,m below $h_0$, and it is cleared when the base is at least 0.5\,m from the near edge of the opening. $B=\mathbf{1}[c_x<-0.1]$ and $S\in\{0,1\}$ indicate a backward and a stop command, respectively (\chapref{subsubsec:episode-commands} and \tabref{tab:common_params}).

\textit{Joint, body, and contact quantities:} $\mathbf{q}$, $\dot{\mathbf{q}}$, and $\ddot{\mathbf{q}}$ are the positions, velocities, and accelerations of the 12 leg joints $j$ (``hips'' denotes the four hip abduction joints), $\mathbf{q}^0$ is the default posture, and $\tau_j$ and $\tau^{\mathrm{c}}_j$ are the applied torque and the PD torque before the limit $\tau_{\max}$; $\dot q_w$ and $\ddot q_w$ are the velocity and acceleration of wheel joint $w$ (wheel radius 0.086\,m). $\mathbf{F}^{\mathrm{s}}_i$ and $\mathbf{F}^{\mathrm{s}}_b$ are the contact-sensor forces of wheel $i$ and of each non-wheel body $b$, which exclude the virtual contact force in \equref{eq:virtual_contact_force}, taken as the maxima over the last three simulation steps; $n_i=1$ if $\lVert\mathbf{F}^{\mathrm{s}}_i\rVert>5$\,N, and $\mathrm{in}_i=1$ if wheel $i$ lies horizontally inside the opening. $h_{\mathrm{b}}$ and $\eta_l$ are the base height and the knee height of leg $l\in\{\mathrm{FR},\mathrm{FL},\mathrm{RR},\mathrm{RL}\}$ (called the elbow for the front legs) above the local terrain surface. $\mathbf{p}_{C}$ approximates the whole-body center of mass, $\bar{\mathbf{p}}_{F}$ is the mean of the wheel positions, $\hat{\mathbf{h}}$ is the horizontal heading, and $y_{\mathrm{F}}$ is the lateral distance from the front-right to the front-left wheel, which is negative when the front legs cross.

\textit{Retreat success and termination:} $\mathrm{succ}$ in \equref{eq:terrain_curriculum} indicates a retreat success (\chapref{subsec:exp-common-settings}) in an episode without failure termination. Failure terminations ($h_{\mathrm{b}}<-0.2$\,m or $-g_z<0.2$) are not bootstrapped, whereas truncations (a return of the base within 0.5\,m of its initial position under the backward command, an exit from the terrain tile, or the 20\,s time limit) are bootstrapped.

\section{Collapse Primitive Details}\label{app:collapse}
This appendix gives the equations of the contact-response law and of the three representative collapse primitives (\chapref{subsec:collapse-primitives}), which realize \(\mathcal{H}\) and \(\Phi\) in \equref{eq:state_multiplexing}. \tabref{tab:common_params} lists the fixed parameters and \tabref{tab:dr} the ranges of the randomized parameters.

\textit{Contact-response law:} Omitting the environment index \(e\), let \(\mathbf{x}_i\), \(\mathbf{v}_i\), and \(\boldsymbol{\omega}_i\) be the position, linear velocity, and angular velocity of wheel \(i\), and \(R\) its radius. Unless lowered by a primitive, the local surface height \(h_i\) equals the nominal height \(h_0\), which includes a static-penetration offset that levels the virtual and surrounding rigid floors under the robot's nominal weight. The penetration \(\delta_i=h_i-(x_{i,z}-R)\) is the depth of the lowest point of the wheel below \(h_i\), and the normal force is the penalty force~\cite{gilardi2002literature,marhefka1999compliant}
\begin{equation}
    f_i^n =
    \alpha_i\,\mathrm{clip}
    \left(
        k\delta_i-2\zeta\sqrt{k m_i}\,v_{i,z}\mathbf{1}[\delta_i>0],
        0,f_{\max}
    \right),
    \label{eq:virtual_normal_force}
\end{equation}
where \(\mathbf{1}[\cdot]\) is the indicator function and \(k\), \(\zeta\), \(m_i\), and \(f_{\max}\) are the contact stiffness, damping ratio, effective mass, and force limit, respectively; the gate \(\mathbf{1}[\delta_i>0]\) removes damping on a wheel above the surface. With the horizontal velocity \(\mathbf{v}_i^t\) of the contact point, i.e., the horizontal component of \(\mathbf{v}_i-R\,\boldsymbol{\omega}_i\times\hat{\mathbf{z}}\), the friction coefficient \(\mu\), and the slip-velocity scale \(\varepsilon\), the regularized Coulomb friction~\cite{castro2020transition}, for which we use a hyperbolic tangent, and the applied force are
\begin{equation}
    \begin{aligned}
        \mathbf{f}_i^t &=
        -\mu f_i^n
        \tanh\!\left(\frac{\lVert\mathbf{v}_i^t\rVert}{\varepsilon}\right)\frac{\mathbf{v}_i^t}{\lVert\mathbf{v}_i^t\rVert},\\
        \mathbf{F}_i &= f_i^n\hat{\mathbf{z}}+\mathbf{f}_i^t+\mathbf{1}[f_i^n>1\,\mathrm{N}]\,\boldsymbol{\sigma}_i ,
    \end{aligned}
    \label{eq:virtual_contact_force}
\end{equation}
where \(\boldsymbol{\sigma}_i\) is nonzero only for Slip.

\textit{Collapse region:} In polar coordinates \((\rho,\vartheta)\) about the region center, which is randomly offset inside the collapse-capable opening, the boundary radius is \(r(\vartheta)=r_0\bigl(1+\mathrm{clip}(\sum_{n=1}^{K}[a_n\cos n\vartheta+b_n\sin n\vartheta],\,-0.7,\,1.6)\bigr)\), with base radius \(r_0\), \(K\) harmonics, and shape coefficients \(a_n\) and \(b_n\); the clip keeps the radius positive. For wheel \(i\) at \((\rho_i,\vartheta_i)\), the membership \(\chi_i=\bigl(1-\tanh\bigl((\rho_i-r(\vartheta_i))/\lambda\bigr)\bigr)/2\), with boundary width \(\lambda\), maps the patch support ratio \(\alpha_{e,p}\in\{0,1\}\) to the wheel as \(\alpha_i=1-(1-\alpha_{e,p})\,\chi_i\), and a wheel with \(\chi_i>1/2\) counts as inside the collapse region. The center offset, \(r_0\), \(a_n\), and \(b_n\) are sampled every episode and stored in \(\mathbf{z}_e\).

\textit{Contact-triggered support loss (Touch):} The patch is triggered when the normal force \(f_i^n\) of a wheel inside the collapse region first exceeds the trigger force \(f_{\mathrm{tr}}\), and it then samples a collapse delay \(\tau_{e,p}\). Its elapsed time \(s_{e,p}\) increases by \(\Delta t\) at every simulation step, and once \(s_{e,p}\) reaches \(\tau_{e,p}\), \(\alpha_{e,p}\) changes from 1 to 0 for the rest of the episode.

\textit{Load-induced subsidence (Sink):} A subsidence amplitude \(A_{e,p}\), which is zero at the start of an episode, lowers the surface under wheel \(i\) to
\begin{equation}
    h_i=h_0-A_{e,p}\max\!\left(1-\frac{\rho_i^2}{r(\vartheta_i)^2},\,0\right)^{2}.
    \label{eq:sink_profile}
\end{equation}
At every simulation step, \(A_{e,p}\) increases by \(\min(\gamma F_{e,p},\dot{A}_{\max})\,\Delta t\), where \(F_{e,p}=\sum_{i}\max(f_i^n-F_y,0)\) is the excess load over the wheels inside the collapse region, \(F_y\) is the yield force, \(\gamma\) is the subsidence gain, and \(\dot{A}_{\max}\) is the maximum subsidence rate. When \(A_{e,p}\) reaches the collapse amplitude \(A_c\), \(\alpha_{e,p}\) is irreversibly set to 0.

\textit{Lateral slip (Slip):} When triggered as in Touch, the patch samples a slip duration \(T\) and a horizontal unit vector \(\mathbf{d}\), approximately orthogonal to the triggering wheel's horizontal velocity on a random side, or in a random direction if this speed is below 0.05\,m/s. At the trigger, the collapse region becomes a depression whose depth increases linearly along \(\mathbf{d}\) up to the depth of the slip depression. While \(s_{e,p}<T\), the region keeps full support (\(\alpha_{e,p}=1\)), and each wheel in contact receives the traction \(\boldsymbol{\sigma}_i=F_s\mathbf{d}\) with nominal slip force \(F_s\), whose magnitude and direction are perturbed at every simulation step. Once \(s_{e,p}\) reaches \(T\), \(\alpha_{e,p}\) changes from 1 to 0 for the rest of the episode.

\FloatBarrier
\bibliographystyle{IEEEtran}
\bibliography{IEEEabrv,bib/mybibfile}

\end{document}